\documentclass[10pt,twocolumn,letterpaper]{article}

\usepackage{cvpr}              % To produce the CAMERA-READY version
\usepackage{graphicx}
\usepackage{amsmath}
\usepackage{amssymb}
\usepackage[numbers,sort]{natbib}
\usepackage{booktabs}
\usepackage{adjustbox}
\usepackage{multirow}
\usepackage{xspace}
\usepackage{xcolor,colortbl}
\usepackage{wasysym}
\usepackage[font=small,labelfont=bf]{caption}
\usepackage{xcolor,colortbl}
\usepackage{verbatim} 
\usepackage{soul} 
\usepackage{amssymb}
\usepackage{shortbold}
\usepackage{comment}
\usepackage[accsupp]{axessibility}

\usepackage[pagebackref,breaklinks,colorlinks]{hyperref}

\usepackage[capitalize]{cleveref}
\crefname{section}{Sec.}{Secs.}
\Crefname{section}{Section}{Sections}
\Crefname{table}{Table}{Tables}
\crefname{table}{Tab.}{Tabs.}

\def\cvprPaperID{4034} % *** Enter the CVPR Paper ID here
\def\confName{CVPR}
\def\confYear{2023}

\newcommand{\nknote}[1]{\textcolor{blue}{NK: #1}}

\definecolor{ibm1}{HTML}{648FFF}
\definecolor{ibm2}{HTML}{DC267F}
\definecolor{ibm3}{HTML}{FE6100}
\definecolor{ibm4}{HTML}{FFB000}
\definecolor{ibm5}{HTML}{785EF0}
\definecolor{ibm6}{HTML}{88CCEE}

\definecolor{ibm_red}{HTML}{E62325}
\definecolor{ibm_green}{HTML}{24A148}
\definecolor{ibm_yellow}{HTML}{f1c21b}
\definecolor{ibm_purple}{HTML}{C22DD5}
\definecolor{ibm_teal}{HTML}{009d9a}
\definecolor{ibm_ultramarine}{HTML}{648FFF}
\definecolor{ibm_blue}{HTML}{002d9c}

\definecolor{dred}{RGB}{242, 220, 219}
\definecolor{dblue}{RGB}{220, 230, 242}
\definecolor{dpurple}{RGB}{235, 212, 225}

\newcolumntype{s}{>{\columncolor{dred}}c}
\newcolumntype{r}{>{\columncolor{dblue}}c}

\newcommand{\urdf}[0]{URDF\xspace} 
\newcommand{\udf}[0]{UDF\xspace}
\newcommand{\mpd}[0]{LDI\xspace}
\newcommand{\occ}[0]{ORF\xspace}
\newcommand{\ours}[0]{D2-DRDF\xspace}

\newcommand{\rayr}{\vec{\rB}}

\newcommand{\imageC}{\textcolor{ibm_green}{Im.}\xspace}
\newcommand{\meshC}{\textcolor{ibm_teal}{M}\xspace}

\newcommand{\oEvent}{\textcolor{ibm_purple}{\mathcal{O}}}
\newcommand{\iEvent}{\textcolor{ibm_ultramarine}{\mathcal{I}}}
\newcommand{\Loo}{\mathcal{L}_{\oEvent\oEvent}}
\newcommand{\Lii}{\mathcal{L}_{\iEvent\iEvent}}
\newcommand{\Lio}{\mathcal{L}_{\iEvent\oEvent}}
\newcommand{\Loi}{\mathcal{L}_{\oEvent\iEvent}}
\newcommand{\Lent}{\mathcal{L}_{\textrm{ent}}}
\newcommand{\Lmind}{\mathcal{L}_{\textrm{sep}}}

\newcommand{\matterport}{Matterport3D\cite{chang2017matterport3d}}

\newcommand{\dur}{d_\textrm{UR}}

\newcommand{\ddr}{d_\textrm{DR}}

\newcommand{\figdir}{./figures/}

\newcommand{\odsDRDF}[0]{ODS DRDF\xspace}
\newcommand{\sprDRDF}[0]{SPR DRDF\xspace}
\newcommand{\suppfigdir}{./supp_figures_lq/}
\newcommand{\parnobf}[1]{\vspace{0mm} \par \noindent {\bf {#1}.}}
\newcommand{\parnoit}[1]{\par \noindent {\it {#1}.}}
\usepackage{amsmath,amsfonts,bm}

\def\eqref#1{equation~\ref{#1}}
\def\1{\bm{1}}

\newcommand{\test}{\mathcal{D_{\mathrm{test}}}}

\DeclareMathAlphabet{\mathsfit}{\encodingdefault}{\sfdefault}{m}{sl}
\SetMathAlphabet{\mathsfit}{bold}{\encodingdefault}{\sfdefault}{bx}{n}

\newcommand{\sigmoid}{\sigma}

\newcommand{\hlc}[2][yellow]{{%
    \colorlet{foo}{#1}%
    \sethlcolor{foo}\hl{#2}}%
}

\begin{document}

% \title{Single Image to 3D from Posed RGB-D}
% \title{Learning 3D from single images using posed data}
% \title{Single image 3D reconstruction from RGB-D scans}
% \title{Single Image 3D reconstruction for novel images using RGB-D scans}
% \title{Generating 3D reconstruction casually }
%Why not RGBDRDF ;)
% \title{Using Depth maps as nuggets of truth to learn Single Image to 3D Reconstruction}
% \title{Tell me what you see? Learning Single Image 3D reconstruction from pose RGBD-Data}
% \title{Implicit scene representation for single images from depth supervision}
% \title{Learning Single image Implicit Scene Representations from Depth Scans}
% \title{Implicit Scene Representations from a single image}
%\title{Single Image Implicit Scene Representation from Depth Scans}
\title{Learning to Predict Scene-Level Implicit 3D from Posed RGBD Data}
\author{Nilesh Kulkarni, Linyi Jin, Justin Johnson, David F. Fouhey\\
University of Michigan\\
% {\tt \small \{nileshk, jinlinyi, justincj, fouhey\}@umich.edu}
}
\maketitle

%%%%%%%%% ABSTRACT
\begin{abstract}
We introduce a method that can learn to predict scene-level implicit functions for 3D reconstruction from posed RGBD data. At test time, our system maps a previously unseen RGB image to a 3D reconstruction of a scene via implicit functions. While implicit functions for 3D reconstruction have often been tied to meshes, we show that we can train one using only a set of posed RGBD images. This setting may help 3D reconstruction unlock the sea of accelerometer+RGBD data that is coming with new phones. Our system, D2-DRDF, can match and sometimes outperform current methods that use mesh supervision and shows better robustness to sparse data. 
\end{abstract}

%%%%%%%%% BODY TEXT

\definecolor{uppink}{HTML}{FA8F8D}
\definecolor{uppurple}{HTML}{7F7FFE}
\definecolor{upgreen}{HTML}{72F4A6}
\section{Introduction}

\begin{figure}
    \centering
    \includegraphics[width=\linewidth]{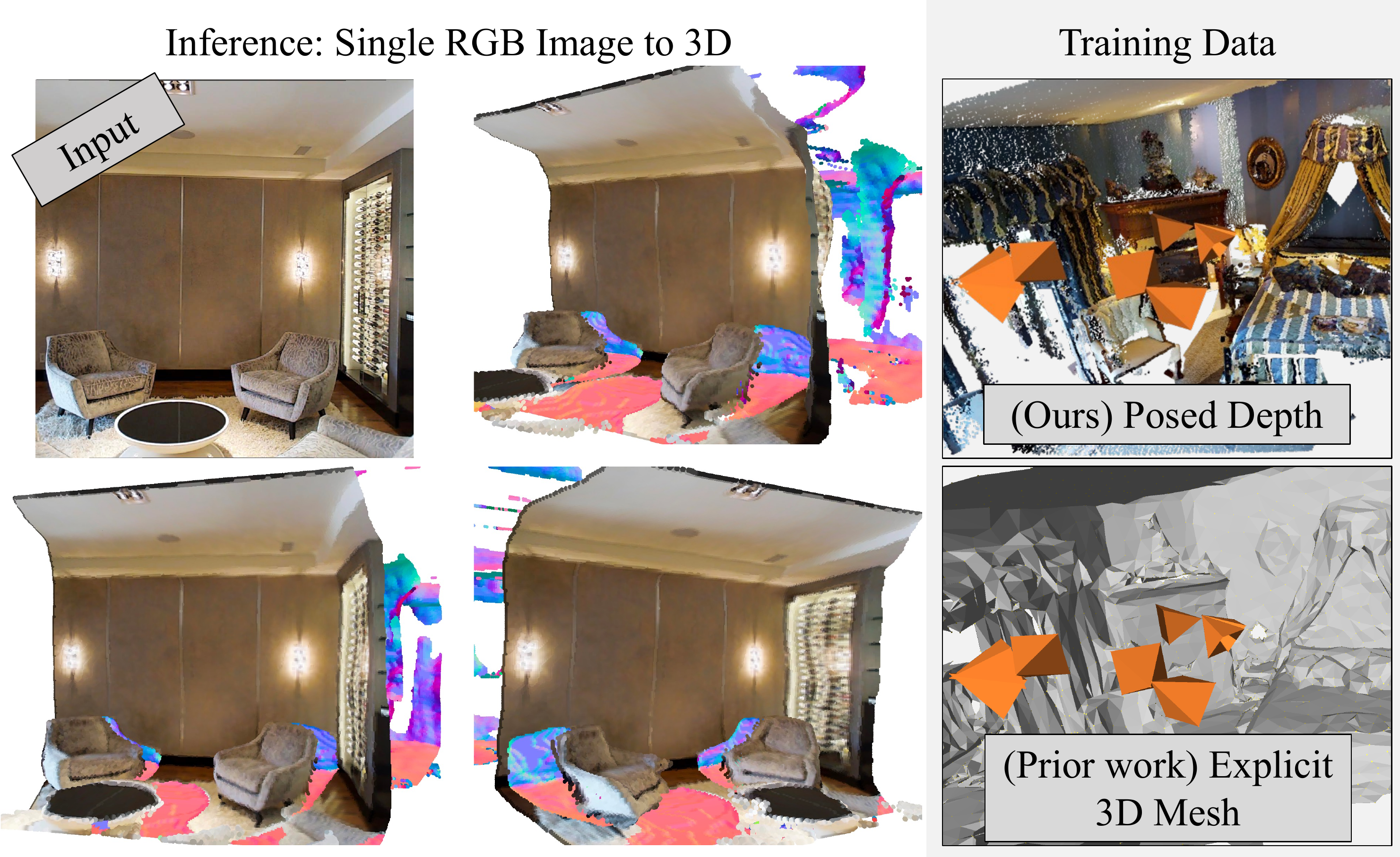}
        \caption{We propose a method, \ours, that reconstructs full 3D from a single RGB image. During inference (left), our method uses implicit functions to reconstruct the complete 3D including visible and occluded surfaces (visualized as surface normals~\includegraphics[width=6pt]{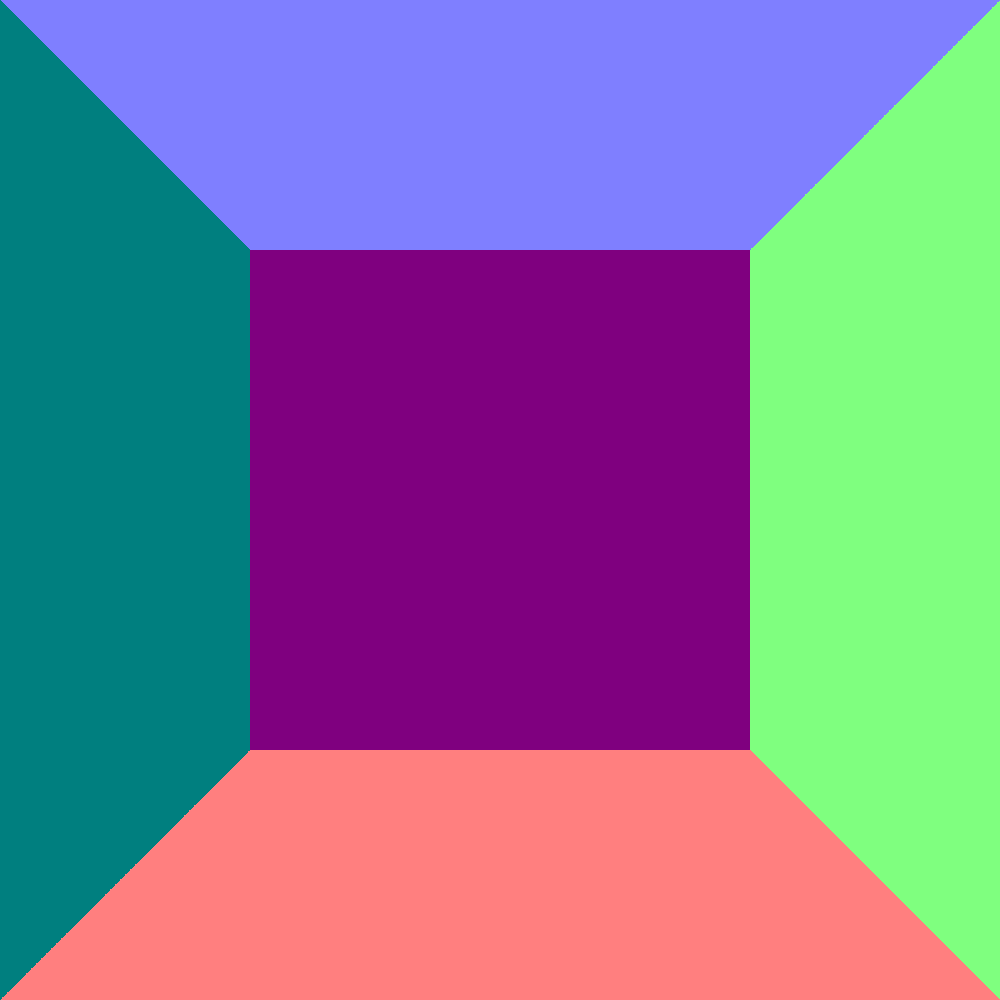}) such as the \hlc[uppurple]{occluded wall} and \hlc[uppink]{empty floor}. For training, our method uses {\it only} RGBD images and poses, unlike most prior works that need an explicit and often watertight 3D mesh.}
        \label{fig:teaser}
    \vspace{-5mm}
    \end{figure}

Consider the image in Figure~\ref{fig:teaser}. From this ordinary RGB image, we can understand the complete 3D geometry of the scene, including the floor and walls behind the chairs. Our goal is to enable computers to recover this geometry from a single RGB image. To this end, we present a method that does so while learning {\it only} on posed RGBD images.

The task of reconstructing the full 3D of a scene from a single previously unseen RGB image has long been known to be challenging. Early work on full 3D relied on voxels~\cite{girdhar2016learning,choy20163d} or meshes~\cite{groueix2018papier}, but these representations fail on scenes due to topology and memory challenges. Implicit functions (or learning to map each point in $\mathbb{R}^3$ to a value like the distance to the nearest surface) have shown substantial promise at overcoming these challenges. When conditioned on an image, these have led to several successful methods.

Unfortunately, the implicit function status quo mainly ties implicit function reconstruction methods to mesh supervision. This symbiosis has emerged since
meshes give excellent direct supervision. However, methods are limited to training with an image-aligned mesh that is usually watertight (and often artist-created)~\cite{mescheder2019occupancy,park2019deepsdf,saito2019pifu} and occasionally non-watertight but professionally-scanned~\cite{kulkarni2022scene,zhu2022differentiable}. 

We present a method, Depth to DRDF (\ours), that breaks the implicit-function/mesh connection and can train an effective single RGB image implicit 3D reconstruction system using a set of RGBD images with known pose. We envision that being able to entirely skip meshing will enable the use of vast quantities of lower-quality data from consumers (e.g., from increasingly common consumer phones with LiDAR scanners and accelerometers) as well as robots. In addition to permitting training, the bypassing of meshing may enable adaptation in a new environment on raw data without needing an expert to ensure mesh acquisition.

Our key insight is that we can use segments of observed free space in depth maps in other views to constrain distance functions. We show this using the Directed Ray Distance Function (DRDF)~\cite{kulkarni2022scene} that has recently shown good performance in 3D reconstruction using implicit functions and has the benefit of not needing watertight meshes for training. Given an input {\it reference} view, the DRDF breaks the problem of predicting the 3D surface into a set of independent 1D distance functions, each along a ray through a pixel in the reference view and accounting for only surfaces on the ray. Rather than use an ordinary unsigned distance function, the DRDF signs the distance using the location of the camera's pinhole and the intersections along the ray. While~\cite{kulkarni2022scene} showed their method could be trained on {\it non}-watertight meshes, their method is still dependent on meshes. In our paper, we show that the DRDF can be cleanly supervised using {\it auxiliary views} of RGBD observations and their poses. We derive constraints on the DRDF that power loss functions for training our system. While the losses on any one image are insufficient to produce a reconstruction, they provide a powerful signal when accumulated across thousands of training views.

We evaluate our method on realistic indoor scene datasets and compare it with methods that train with full mesh supervision.  Our method is competitive and sometimes even better compared to the best-performing mesh-supervised method~\cite{kulkarni20193d} with full, professional captures. As we degrade scan completeness, our method largely maintains its performance while mesh-based methods perform substantially worse. We conclude by showing that fine-tuning of our method on a handful of images enables a simple, effective method for fusing and completing scenes from a handful of RGBD images.

\section{Related Work}
\label{sec:related}

Our approach infers a complete 3D scene from a single RGB image using implicit functions that are supervised via posed RGBD scans. This touches on several long-term goals of 3D computer vision that we discuss below.

\parnobf{Reconstructing Scenes from a Single Image} At test time, our system produces a {\it full} 3D reconstruction from a single RGB image, including occluded regions. This means that our desired output goes beyond 2.5D properties such as depth~\cite{Chen2016SingleImageDP, Eigen2015PredictingDS}, surface normals~\cite{fouhey2013data,Wang15,Eigen2015PredictingDS}, or other intrinsic-image properties~\cite{shi2017learning, janner2017self}. Works on attempting to reconstruct complete 3D usually have focused on reconstructing objects from image by using voxels~\cite{girdhar2016learning, choy20163d} or meshes~\cite{gkioxari2019mesh, groueix2018papier}, point clouds~\cite{lin2018learning, fan2017point} or CAD models~\cite{izadinia2017im2cad}. These are usually trained on synthetic datasets like ShapeNet~\cite{chang2015shapenet} or scene datasets~\cite{song2016ssc, fu20213d, roberts:2021} and do not generalize well to realistic scenes. 
Another line of works tries to learn 
holistic structures~\cite{nie2020total3dunderstanding, zhang2021holistic} or planar surfaces~\cite{liu2019planercnn, yu2019single, jiang2020peek} from realistic scanned mesh data~\cite{dai2017scannet, chang2017matterport3d}  Creating realistic mesh based data for scenes requires post-processing using Poisson surface reconstruction~\cite{kazhdan2006poisson, dai2017bundlefusion} which leads to deviation from raw captures. Manually aligning them is expensive  hence 3D object-aligned datasets like~\cite{song2015sun, sun2018pix3d} are scarce and limited in diversity. Our method avoids these limitations by directly operating on the raw captured RGBD data.

\parnobf{Reconstructing Scenes from Posed Scans}
There has been considerable work on using multiview RGBD data at inference time to produce 3D reconstructions, starting with analytic techniques~\cite{curless1996volumetric,kutulakos2000theory,dai2017bundlefusion,izadi2011kinectfusion} and now using learning~\cite{sun2021neuralrecon,wang2018mvdepthnet,huang2018deepmvs}. We use some similar tools to this community: for instance, the ray distances we use are known in this community as a projected distance functions~\cite{curless1996volumetric}. However, their works solve a fundamentally different problem by using posed RGBD data at test time: their goal is to produce a reconstruction from a set of posed RGBD images from one particular scene; our goal is to use a large dataset of posed RGBD data to train a neural network that can map a new single RGB image to a 3D reconstruction. 

Our goal of learning to predict reconstructions from a single previously unseen image also distinguishes our work from NeRF~\cite{mildenhall2020nerf} and other similar radiance field approaches~\cite{chen2022tensorf}. Their goal is to learn a radiance field for a {\it particular} scene from a set of posed scans. There are methods that try to predict this radiance field from a single image~\cite{yu2020pixelnerf}; we compare with a model using similar losses and find that an objective specialized for 3D works better.

\parnobf{Implicit Functions for 3D Reconstruction} We perform reconstruction with implicit functions. Implicit functions have used for shape and scene modeling as level sets~\cite{malladi1995shape}, signed distance functions~\cite{park2019deepsdf}, occupancy function~\cite{saito2019pifu, mescheder2019occupancy, peng2020convolutional}, distance functions~\cite{kulkarni2022scene, sitzmann2020implicit, chibane2020ndf} and other modifications like ~\cite{ye2022gifs}. Our work has two key distinctions. First, many approaches~\cite{sitzmann2020implicit,chibane2020ndf,ye2022gifs} {\it fit} an implicit function to one shape (i.e., there is no generalization to new shapes). In contrast, our work produces reconstructions from previously unseen RGB images. Second, the methods that predict an implicit function from a new image~\cite{kulkarni2022scene,zhu2022differentiable} assume access to a mesh at training time. On other hand, our work assumes access only to posed RGBD data for training.

The most similar work to ours is~\cite{kulkarni2022scene} that also aims to learn to reconstruct scenes from a single image by training on realistic data. While~\cite{kulkarni2022scene} shows how to reduce supervision requirements by enabling the use of non-watertight meshes, their method is still limited to mesh supervision. We show how to use posed RGBD data for supervision instead of using an image-aligned mesh. This substantially reduces the requirements for collecting training data.

\begin{comment}
\parnobf{Single Image to 3D}. Our work is closest to work in this area, blah blah.
\nknote{see factored3d, meshrcnn, georgia paper, planercn, etc}
\parnobf{3D Reconstruction using Implicit Functions}
\nknote{lots of implicit work}
\parnobf{3D Reconstruction beyond visible regions}
\nknote{see drdf,  pee ka boo}
\end{comment}
\section{Pixel-Aligned 3D Reconstruction \& DRDF}

\begin{figure}[t]
    \includegraphics[width=0.95\linewidth]{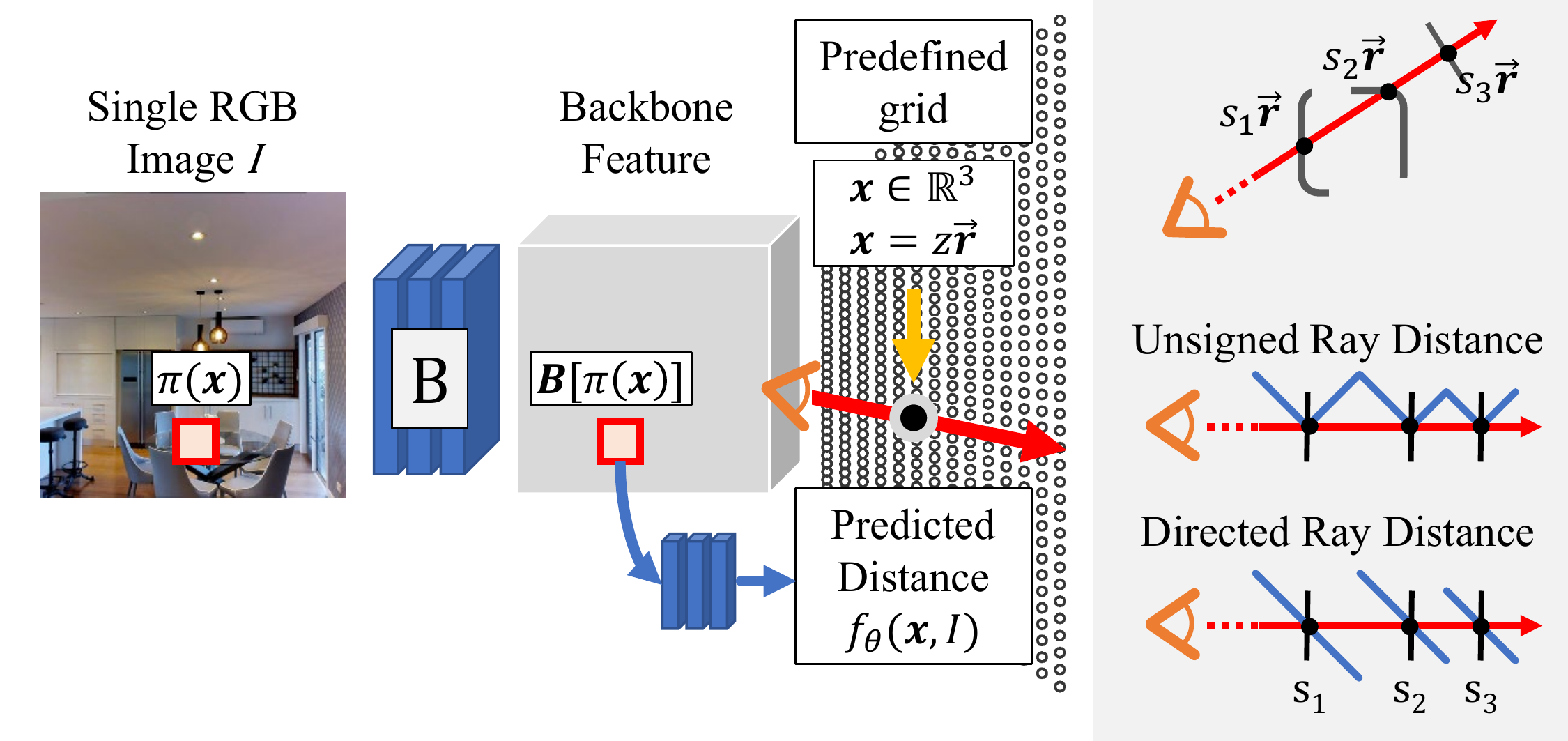}
    
     \caption{{\bf Model and DRDF overview.} (Left) Given a single input RGB image $I$ and query point $\xB$ on a ray $\rayr$ in a pre-defined grid, all models, like~\cite{yu2020pixelnerf} extract image features at the projection $\pi(\xB)$ of $\xB$ with a backbone and then predict the distance using a MLP on the backbone features and a positional encoding of $\xB$. (Right) Given a ray $\rayr$ through the scene, ray distance functions are defined as the distance to intersections of the ray with the scene. This permits analyzing the 3D scene as a set of 1D distances per ray. The Directed Ray Distance Function is further signed to be positive before and negative after an intersection.
     }
     \label{fig:model}
     \vspace{-5mm}
\end{figure}

We propose a method to predict full 3D scene structure from a previously unseen RGB image. At inference time, our method receives a single image with no depth information. We refer to this view as the {\it reference view}. As output, the method produces an estimate of a {\it distance function} at each point in a pre-defined set of 3D points. Given this volume of predicted distances, one can {\it decode} the predicted distance into a set of surfaces: e.g., if the predicted distance were the unsigned distance to the nearest surface, one could declare all points with sufficiently small predicted distance to be surfaces. At training time, our network is given supervision for predicting its distance function. Previously, supervision been done via a mesh that provides oracle distance calculations. In \S\ref{sec:method}, we show how to derive supervision from posed RGBD images at {\it auxiliary views} instead.

The approaches in the paper follow pixelNerf~\cite{yu2020pixelnerf} for predicting a distance at a 3D point $\xB$. The network accepts an image $I$ and 3D point $\xB$ and makes a prediction $f_\theta(\xB,I)$. As shown in Fig.~\ref{fig:model}, the network extracts a convolutional feature from a backbone $\BB$ at the projection of $\xB$, denoted $\pi(\xB)$. The image feature is concatenated to a positional encoding of $\xB$ and put into a MLP to produce the final prediction $f_\theta(\xB,I)$. By repeating this inference for a set of 3D points, the network can infer the full 3D of the scene.

\label{sec:prelim}

\parnobf{Training} In a conventional setup, the network $f_\theta(\xB,I)$ is trained by obtaining samples from a mesh and directly training the network via regression. Given a mesh that is co-aligned with an image, one can obtain an $n$ tuples consisting of an image $I_i$, 3D point $\xB_i$ and ground-truth distance $d_i$. Then the network is trained by minimizing the empirical risk $\frac{1}{n} \sum_{i=1}^n \mathcal{L}(f_\theta(\xB_i,I_i), d_i)$ for some loss $\mathcal{L}$ like L1.

\par \noindent {\bf What Distance Function Should We Predict?} There are multiple distance functions in the literature, and the above formulation enables predicting any of them.
To explain distinctions, assume we are given a point $\xB$ at which we want to evaluate a distance and a scene $\mathcal{S} \subseteq \mathbb{R}^3$ that we want to evaluate the distance to.

In a standard distance function, one defines a distance from a point $\xB$ to the {\it entire} scene $\mathcal{S}$. For instance, the {\it Unsigned Distance Function} (UDF) is defined as $\min_{\xB' \in \mathcal{S}} ||\xB - \xB'||$. In a {\it ray-based distance function}~\cite{kulkarni2022scene,curless1996volumetric}, the distance is defined from $\xB$ only to the points in the scene that lie on the ray $\rayr$ from the reference view pinhole towards $\xB$. In other words, the ray distance function is defined as the distance to the intersections of the ray with the scene. For instance, the {\it Unsigned Ray Distance Function} (URDF) is defined as $\min_{\xB' \in \mathcal{S}, \xB'~\textrm{on}~\rayr} ||\xB - \xB'||$. This definition means that when predicting the distance at a point $\xB$, all of the points that define the ray distance function project to the same pixel, $\pi(\xB)$, which in turn means that neural networks do not need as large receptive fields~\cite{kulkarni2022scene}.

Since a ray-based distance function is defined {\it only} by scene points that intersect a ray, we can reduce the 3D problem to a 1D problem, as illustrated in Fig.~\ref{fig:model}. Rather than represent each 3D point as a 3D point, we represent it as the distance along the ray $\rayr$: the vector $\xB$ is represented via a scalar $z$ such that $\xB = z \rayr$. Then, if the $i^{th}$ intersection along the ray is $s_i \rayr$ for $s_i \in \mathbb{R}$, the URDF is defined as the distance to the nearest intersection along that ray, or $\dur(z) = \min_{i=1, \ldots, k} |z-s_i|$. 

\parnobf{Directed Ray Distance Function} The Directed Ray Distance Function (DRDF) is a ray distance function that incorporates a sign into the URDF. In particular, the DRDF is defined as $\ddr(z) = \textrm{dir}(z) \dur(z)$, where the predicate $\textrm{dir}(z)$ is positive if the nearest intersection is ahead of $z$ along the ray from the pinhole and negative otherwise. This is pictured in Fig.~\ref{fig:model}. The presence of both positive and negative components leads to better behavior under uncertainty~\cite{kulkarni2022scene}. 

\parnoit{Critical Properties of DRDFs} Two critical properties that we use  to define supervision are that at $z$, the distance to the nearest intersection is $|\ddr(z)|$ and the nearest intersection on the ray is at $z+\ddr(z)$.

\section{Depth-Supervised DRDF Reconstuction}
\label{sec:method}

Having introduced the setup in \S\ref{sec:prelim}, we now explain how to train a network to predict an image-conditioned DRDF {\it without} access to an underlying mesh but instead access to posed RGBD images. Concretely, given a point $z \rayr$, we aim to define a loss function $\mathcal{L}$ that evaluates a network's prediction of the DRDF $f_\theta((z \rayr), I)$.

\begin{figure}[t]
    \centering
    \includegraphics[width=0.95\linewidth]{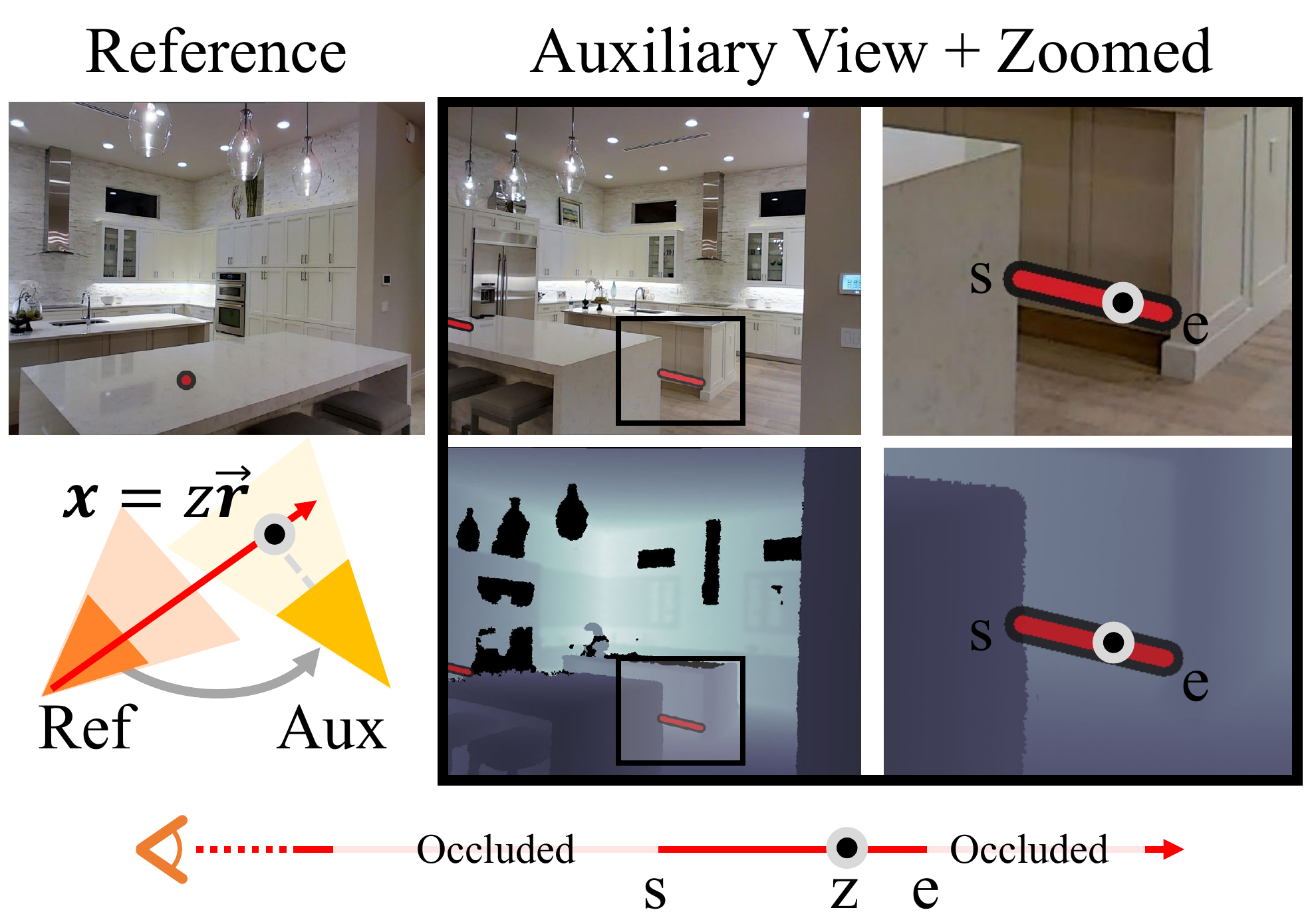}
    \vspace{-4mm}
    \caption{A \textcolor{ibm_red}{red ray} originates in the reference camera. Given a point $\xB$ that is $z$ units along the ray \textcolor{red}{$\rayr$}, we can test if $z\rayr$ is visible in an auxiliary view by comparing its projected depth with the RGBD image. The point $z$ is visible, along with a segment of the ray from $s$ to $e$ units. Since the distance to the nearest intersection is $|\ddr(z)|$ and we know there are no intersections within $(e-z)$ units of $z$, we can constrain that $|\ddr(z)| \ge (e-z)$.} 
    \label{fig:ray_setup}
    \vspace{-4mm}
\end{figure}

Our key insight is that we can {\it sometimes} see this point $z \rayr$ in other views and these observations of $z \rayr$ provide a constraints on what the value of the DRDF can be. Consider, for instance, the point $3$m out from reference view camera pinhole along the red ray in Fig.~\ref{fig:ray_setup}. If we project this point into the auxiliary view, we know that its distance to the camera is closer to auxiliary view's pinhole than the depthmap observed at the auxiliary view. Moreover, we can see that a segment of the ray is visible, starting at the point $s \rayr$ and ending at the point $e \rayr$ with the ending point $e \rayr$ closer to $z \rayr$ than $s \rayr$ is.

We can derive supervision on $\ddr(z)$ from the segment of visible free space observable in the auxiliary view by using the fact that the nearest intersection at $z$ is $|\ddr(z)|$ units away. Since $e$ is nearer than $s$, and there are {\it no} intersections in the free space between $s$ and $e$, we know that there cannot be an intersection within $(e-z)$ of $z$. This gives an inequality that $|\ddr(z)| \ge (e-z)$: otherwise there would be an intersection in the visible free space. If the ends of the segment are actual intersections where the ray visibly intersects an object, we obtain an equality constraint on $\ddr(z)$ since we actually see the nearest intersection. In other cases, the ray simply disoccludes or is occluded, and we can only have an inequality. We denote intersection start/end events as $\iEvent$ and occlusion start/end events as $\oEvent$ and distinguish them by checking if the projected z-value of the ray in auxiliary view is sufficiently close to the recorded auxiliary view's depthmap value.

\newcommand{\first}{F}
\newcommand{\slf}{l_s}
\newcommand{\elf}{l_e}

\subsection{Losses}
\label{sec:method_losses}

\begin{figure*}
    \centering
    \begin{adjustbox}{max width=\linewidth}
    \begin{tabular}{c@{\hskip4pt}c@{\hskip1pt}c@{\hskip1pt}c}
    \noindent\includegraphics[width=0.365\linewidth,trim={0 0cm 0 0},clip]{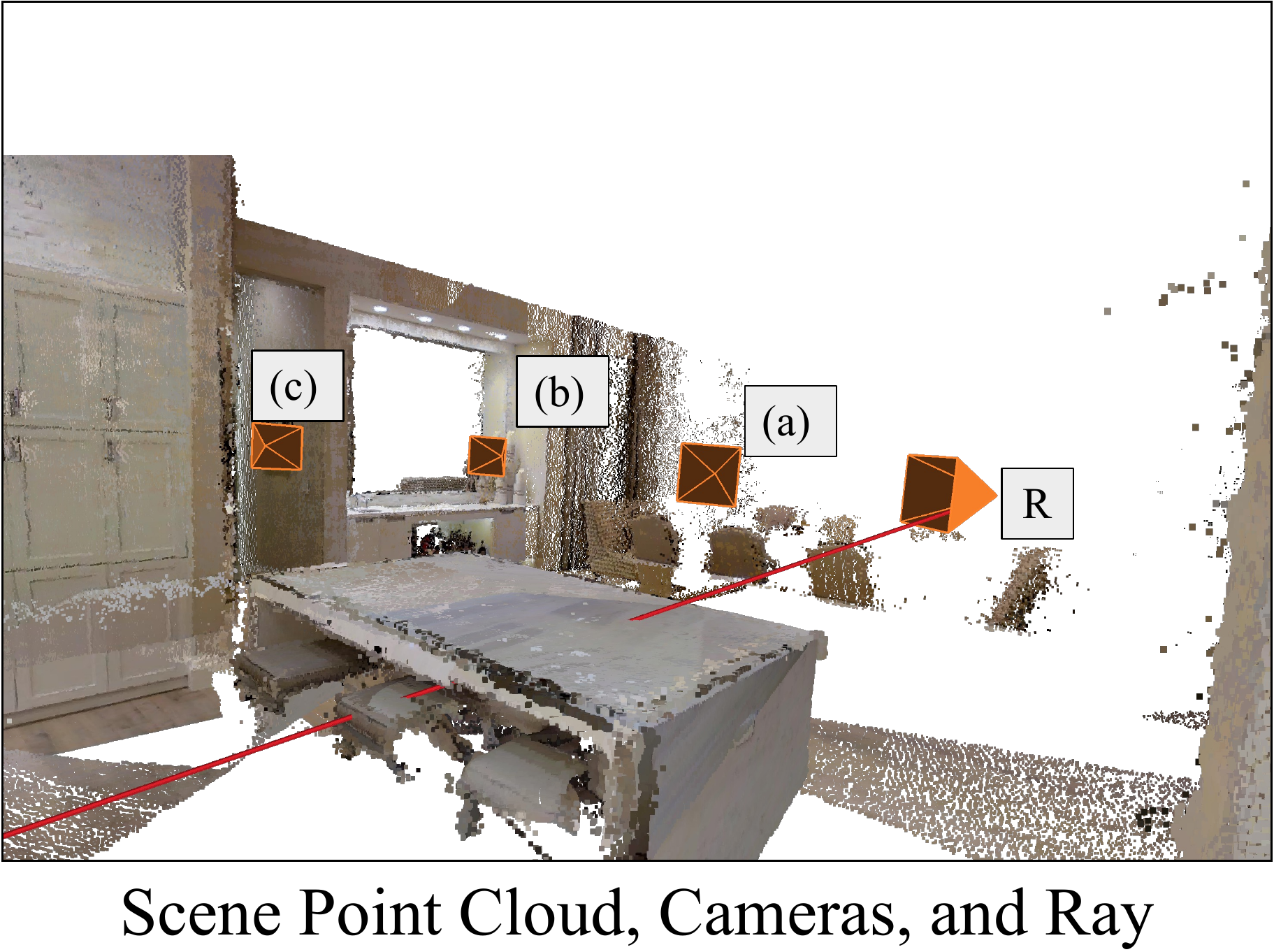} &
    \noindent\includegraphics[width=0.62\linewidth, trim={0 0cm 0 0},clip]{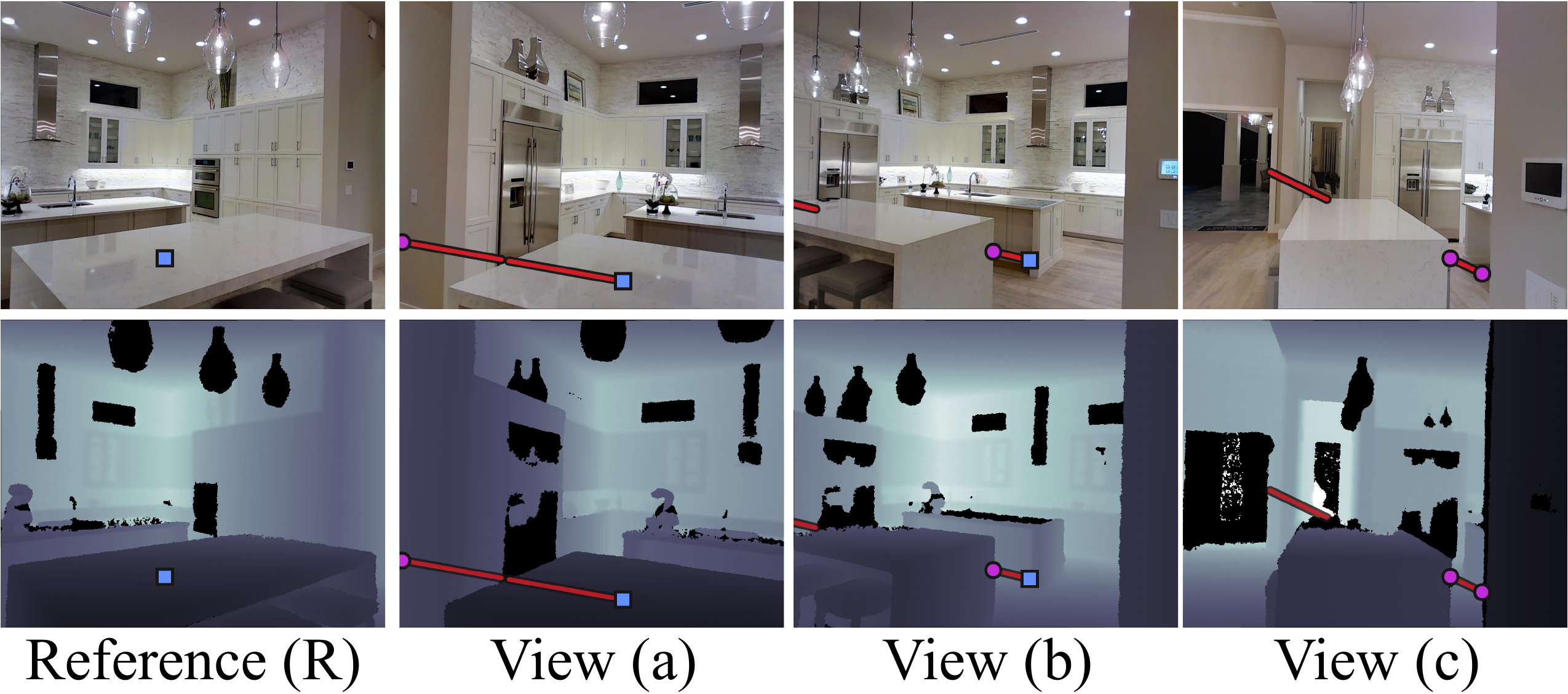}
    \vspace{-4mm}
    \end{tabular}
    \end{adjustbox}
    \caption{A \textcolor{ibm_red}{red ray} originates from the reference camera and intersects the table before entering the island. This ray is seen by three other {\it auxiliary} views (a, b, c). For each view we show just one of the many segments in that view for readability, showing intersections as \textcolor{ibm_ultramarine}{blue {$\blacksquare$}s} and occlusions as \textcolor{ibm_purple}{purple {\newmoon}s}. The segments in views (a) and (b) are of the form $\oEvent\iEvent$, where the $\oEvent$ comes from the ray entering the frustum in (a) and disoccluding in (b). In view (c), we show an $\oEvent\oEvent$ segment. We discuss the penalties for these segments in \S\ref{sec:method_losses}
    }
    %\caption{A \textcolor{ibm_red}{red ray} originating from the camera in reference view and intersects the bar counter (at the \textcolor{ibm_ultramarine}{blue $\blacksquare$}) has a first \textcolor{ibm_ultramarine}{intersection event} visible in (a). This ray then intersects with the  bar stools tucked underneath the counter as seen from (d), and the kitchen island as seen from (b).  The ray has a $\oEvent\oEvent$ segment shown by \textcolor{ibm_purple}{purple occlusion events} in (c), denoted by \textcolor{ibm_purple}{\newmoon}. Here the ray does not intersect any geometry creating inferred from depth map (c).  We can see how the captured depth maps have missing values which leads to unknown information along the ray. For all the auxiliary views we only show some of the segments from the complete segment list to ease readability.}
    \label{fig:ray_events}
    \vspace{-4mm}
\end{figure*}

We now convert the concept of using freespace to constrain the DRDF into concrete losses. Recall that our goal is to evaluate a prediction $y = f_\theta(z \rayr,I)$ from our network and that the point at $z$ along $\rayr$ is in some segment of visible freespace from $s$ to $e$ along $\rayr$. Our goal is to penalize the prediction $y$. Since there are two start/end event classes ($\iEvent$, $\oEvent$), there are four segment types: $\iEvent\iEvent$, $\iEvent\oEvent$, $\oEvent\iEvent$, and $\oEvent\oEvent$. We show several of these segments in Figure~\ref{fig:ray_events}.

\begin{figure}[t]
    \centering
    \includegraphics[width=\linewidth]{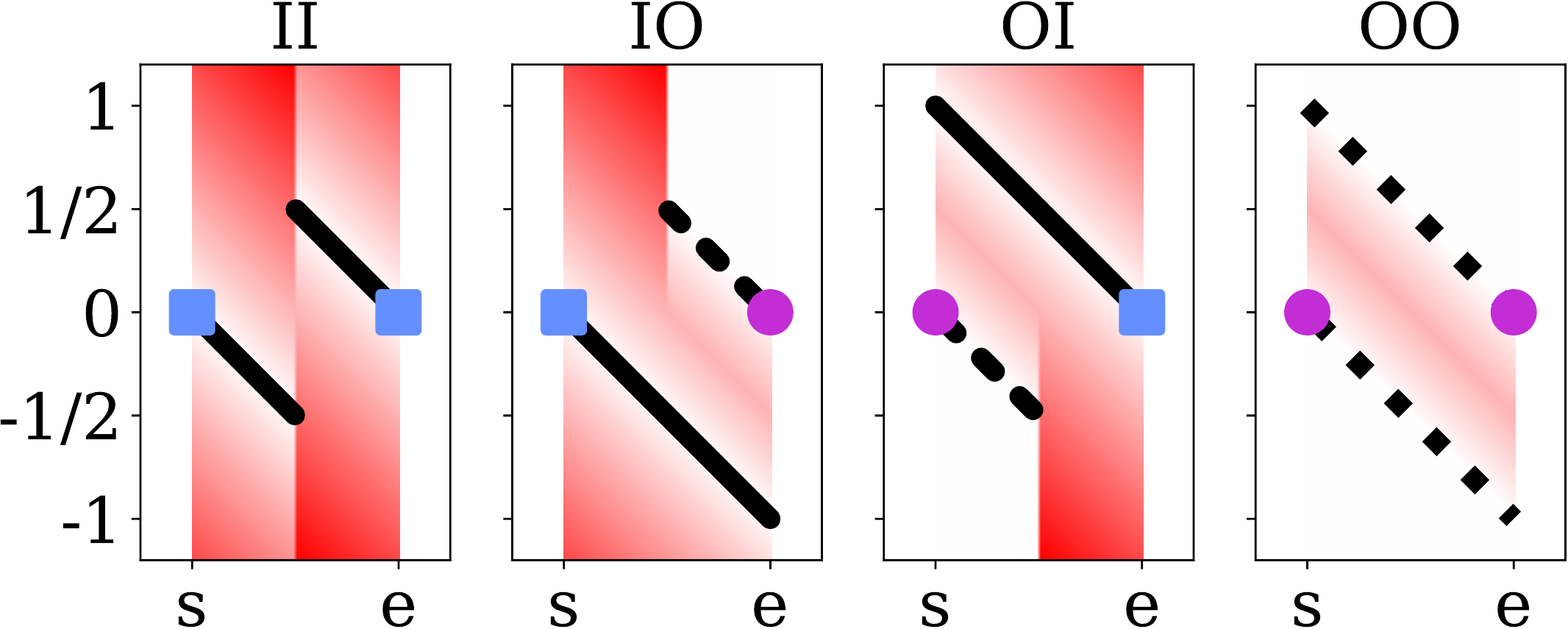}
        \caption{Loss functions for a starting $s$ and ending $e$ events spaced a unit apart. Intersection events (\textcolor{ibm_ultramarine}{blue $\blacksquare$}) define the DRDF precisely, and we penalize the network from deviating. Occlusion events (\textcolor{ibm_purple}{\bf purple \newmoon}) provide bounds that we penalize the network for violating. We plot $\slf$ as a line from the start event and $\elf$ as a line from the end event.        
        Loss: 0     \includegraphics[width=30pt,height=6pt]{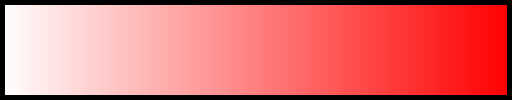} 1.7   
        }
        \label{fig:losses}
        \vspace{-4mm}
    \end{figure}

To assist the loss definitions, we define variables $\slf = s-z$, and $\elf = e-z$ which we plot  in  Fig.~\ref{fig:losses} as a function of $z$. $\slf$  is the value of $\ddr$ if the closest intersection is at $s$, and $\elf$ is the value of $\ddr$ if the closest intersection is at $e$. The values $\slf$, $\elf$ define equalities for intersections and inequalities for occlusions. % and are plotted as a function of $z$ in Fig.~\ref{fig:losses}.

\parnobf{$\Lii$: $\iEvent\iEvent$ segment} 
Given a segment bounded by two intersections, the nearest intersections are known exactly as $\slf$ or $\elf$ depending on whether $z < \frac{s+e}{2}$ or not.  We penalize the $\ell_1$ error between the prediction $y$ and the known DRDF, or $\Lii(y) = |y - \slf|$ if $z < \frac{s+e}{2} $ and $|y - \elf|$ otherwise. This penalty is is zero only when $y$ is equal to the known DRDF.

\parnobf{$\Loo$: $\oEvent\oEvent$ segment}
Given a segment bounded by two occlusion events, the exact DRDF is not known, but the visible free space rules out potential values. Since $|\ddr(z)|$ is the distance to the nearest intersection, $\ddr(z)$ cannot lie in $[\slf,\elf]$ since such a value would imply an intersection in free space between $s$ and $e$. We penalize incursions into $[\slf,\elf]$ with a $\ell_1$ penalty: if we denote halfway between $\slf$ and $\elf$ as $h$, then this can be done as $\Loo(y) = \max(0, \elf - h - \left| y - h\right|)$. The resulting penalty is zero if $y \le \slf$ or if $y \ge \elf$.

%Depending on whether the closest intersection to $x$ is before or after $x$, we obtain different constraints. If the nearest intersection is before $z$, then the start of the visible segment $s$ gives an upper bound on how close that intersection is. This gives an upper bound in case $s(z)$ is negative, namely that $y(z) \le u$ where $u(z)=(s-z)$: if $y(z) > u$, then this would imply the existence of an intersection in the visible portion. Similar reasoning shows that if $y(z) > 0$, then we obtain a lower bound for when $y(z) > 0$ of the form $l(z) = (e-z)$.

%With two occlusion events, it is not clear which bound applies, but we know that $d(x)$ {\it cannot} be between $u$ and $l$. We linearly penalize violations of this constraint via $\mathcal{L}_{OO} = \max(0,(l(z)-u(z))/2 - |y(z)-(u(z)+l(z))/2|)$.

\parnobf{$\Lio$: $\iEvent\oEvent$ segment}
When the segment is bounded by an intersection event followed by an occlusion event, the situation is more complex and we define $\Lio$ piecewise. In the first half of the segment($z < \frac{s+e}{2}$), $\ddr$ is exactly known, and so we can use an $\ell_1$ penalty like the $\iEvent\iEvent$ case, so $\Lio(y) = |y-\slf|$. In the second half, there are two options. If the nearest intersection is $s$, then $\ddr(z) = \slf$. Otherwise, the nearest intersection is unknown but after $e$ and so $\ddr(z) > \elf$ must hold (since $\ddr(z) \le \elf$ would imply the existence of an intersection before $e$). We take the minimum of errors for the two cases: $\ell_1$ distance to $\slf$ and a $\ell_1$ penalty function $\max(0,\elf - y)$, resulting in
$\Lio(y) = \min(\max(0,\elf-y),|y-\slf|)$. This part of the penalty is zero if either $y = \slf$ or if $y > \elf$.

\parnobf{$\Loi$: $\oEvent\iEvent$ segment}
The $\Loi$ loss is defined symmetrically to $\Lio$, simply by exchanging the role of $s$ and $e$. In addition to occluded regions, this loss occurs in the reference view up to the depthmap, where a disocclusion into the reference camera's view is followed by an intersection.

To assist the network, we add two auxiliary losses that are true statistically: $\Lmind$ represents a prior that surfaces tend to be separated by distances and $\Lent$ captures a property of the DRDF that is true in the limit if our observations are randomly chosen. 

\parnobf{$\Lmind$: Minimum Separation Loss} Since the cameras never sees the insides of objects, there is no incentive to predict inside objects. This prevents the generation of zero crossings, e.g., after the first intersection. To assist the network, we add a loss $\Lmind$ that assumes that surfaces are separated by a minimum distance unless there is evidence otherwise. We continue the DRDF's known value for $t$ = $0.2m$ before and after each intersection event to make a continued value $c$. We then penalize  $\Lmind(y) = |y-c|$, so long as there is no conflicting free space evidence from another view. 

\parnobf{$\Lent$: Sign Entropy Loss} The occlusion-based constraints can be satisfied by making $y$ positive or negative. In theory, after the first intersection, $\ddr$ is positive half of the time. Learning to produce the signs uniformly with gradient descent is difficult due to the large loss between the acceptable solutions. To encourage a uniform distribution of signs, we would like to maximize the entropy of the distribution of the signs of the predictions (achieved when the distribution is uniform). Since sign is non-differentiable, we optimize a differentiable surrogate. Suppose $Y$ is the set of predictions at points that are occluded in the reference view and $H$ is the binary entropy $H(p) = p \log(p) + (1{-}p) \log(1{-}p)$. Then our loss is
$\Lent = H(\sum_{y \in Y}\sigmoid(y/\tau)/|Y|)$ where $\sigmoid$ functions like a soft-sign. As seen in the supplement, $\Lent$ can be minimized by distributing $Y$ symmetric about $0$. Using $\Lent$ improves the prediction and Scene F1 by 2 points.

\subsection{Implementation Details}

\parnobf{View Selection} For every candidate auxiliary view we compute the fraction of visible points in this view that are occluded in the reference view. Auxiliary views with large number of such points provide supervision for key occluded regions in the reference view. We sample up to $20$ auxiliary views per reference image from this set of views.

\parnobf{Sampling Strategy} Given a set of fixed auxiliary views and a reference view, we sample over $200$ rays per input image with $512$ points per ray. We re-balance this set by sampling $20K$ points that are visible and $20K$ points that are occluded in reference view. We rebalance as most points are from the region between the camera and the first hit.

\parnobf{Combining segments from different views} We merge information from multiple posed RGBD images to produce a concise merged set of non-overlapping segments along the ray. This prevents double-counting losses (e.g., if a region of the ray is seen by multiple auxiliary views). We safely merge segments that provide the same information: e.g., if one depthmap provides an $\oEvent \oEvent$ segment that is  contained within another depthmap's $\iEvent \iEvent$ segment, then the $\oEvent \oEvent$ segment can be safely dropped since $\Loo(y) \le \Lii(y)$. When segments disagree (e.g., due to inaccurate poses), we keep the segment with more auxiliary views in agreement. This approach handles merging $\Lmind$: $\Lmind$ is seen by no auxiliary views, so any visible freespace overrules it.

\parnobf{Network Architecture} We follow~\cite{kulkarni2022scene} to facilitate fair comparison. Additionally, we clamp the outputs of our network to be $\in [-1, 1]$ by applying a $\tanh$ activation and adjust the loss to account for this clipping.

%\dfnote{Need to mention clipping and tanh if those are still happening}

\parnobf{Training} We follow a two-stage training procedure. We first train with only the reference view followed by adding auxiliary views. In the first stage, we train for $100$ epochs minimizing $\Loi$ and  $\Lmind$. We then train for $100$ epochs with auxiliary losses, minimizing a sum of the segment losses $\Lii$, $\Lio$, $\Loi$, $\Loo$, and $\Lmind$ as well as $\lambda \Lent$ with $\lambda$ set to $0.1$ to balance loss scales. We minimize the loss with AdamW~\cite{loshchilov2017decoupled, kingma2014adam} as the optimizer with learning rate warmup for 0.5\% of the iterations followed by cosine learning rate decay with maximum value $3 \times 10^{-4}$. Our models are implemented using PyTorch~\cite{NEURIPS2019_9015} and visualizations in this paper use PyTorch3D~\cite{ravi2020pytorch3d, lassner2020pulsar}.

\section{Experiments}
\begin{figure*}[t]
\centering
    \includegraphics[width=\linewidth]{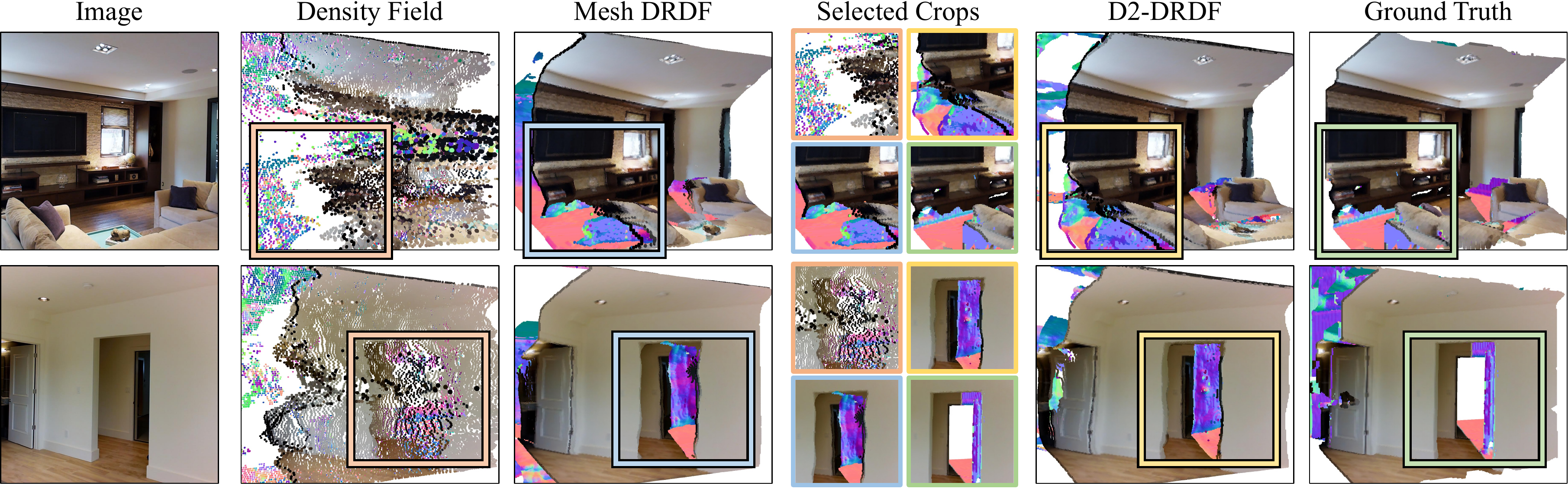}
    \caption{{\bf Comparison with baselines.} 3D outputs generated by all methods trained on Matterport3D. We color the first intersection with image colors and occluded intersections with computed surface normals. We highlight regions of interest in the reconstructions in selected crops. \ours achieves results on par with Mesh DRDF while the density fields baselines fails to model the occluded parts faithfully. In row 1, our method recovers the back of sofa, and a hidden room behind the hallway in row 2. Surface Normal Map \includegraphics[width=6pt]{figures/normal_map2.png}} 
    \label{fig:comparison}
    \vspace{-5mm}
\end{figure*}

We evaluate our method to address three experimental questions. First, we examine how training with RGBD data compares with mesh supervision. Next, we test how RGBD 
and mesh supervision compare when  one has less complete scans. Finally, we show that our method can quickly adapt to multiple posed RGBD inputs.  

\parnobf{Metrics} Throughout, we follow~\cite{kulkarni2022scene} and  use two metrics that evaluate predictions against a ground-truth mesh. Following~\cite{seitz2006comparison,tatarchenko2019single}, these metrics are based on:  Accuracy/Acc (the fraction of predicted points that are within $t$ to a ground-truth point), Completeness/Cmp (the fraction of ground-truth points that are within $t$ to a predicted point), as well as F1 (the harmonic mean of Accuracy and Completeness). $t$ for both Acc and Cmp is 0.5m. In {\bf (Scene Acc/Cmp/F1)}, we evaluate the predicted mesh of the full scene against the ground-truth, using 10K samples per mesh. In {\bf (Ray Occ.~Acc/Cmp/F1)}, We evaluate the performance on occluded points, evaluating per-ray and then averaging. We define occluded points for both the ground truth and prediction as any surface past the first intersection. Ray Occ.~is a challenging metric as mistakes in one ray cannot be accounted for in another ray. 

\parnobf{Datasets}
We use Matterport3D~\cite{chang2017matterport3d} as our primary dataset  following an identical setup to~\cite{kulkarni20193d}. We choose Matterport3D because it has substantial occluded geometry (unlike ScanNet~\cite{dai2017scannet}) and was captured by a real scanner (unlike 3DFront~\cite{fu20203dfront}, which is synthetic). We note that we use the {\it raw images} captured by the scanner rather than re-renders. We follow the split from~\cite{kulkarni2022scene}, which splits train/val/test by house into 60/15/15 houses. After filtering and selecting images, there are 15K/1K/1K input images for each split.

\begin{figure}[t]
\centering
    \includegraphics[width=\linewidth]{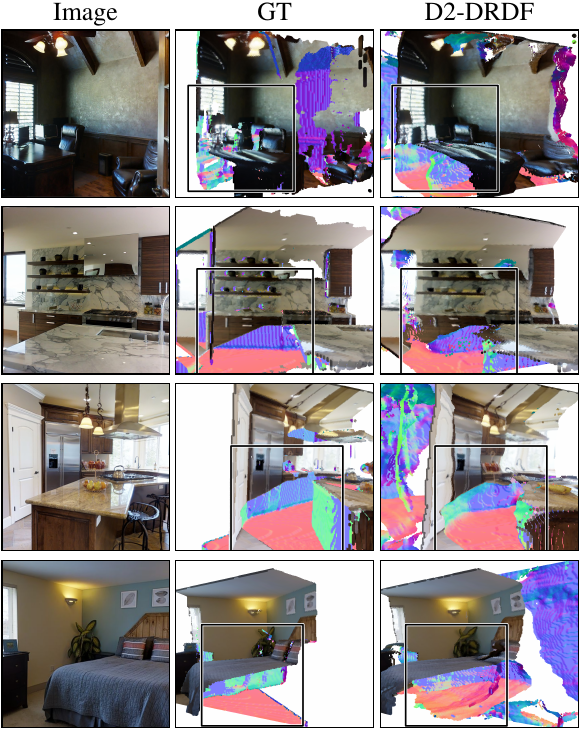}
    \caption{{Novels Views} Comparison between \ours and Ground Truth(GT) from novel views. Rows 1, 2 are from unseen images on Matterport3D~\cite{chang2017matterport3d} and 3,4 from Omnidata~\cite{eftekhar2021omnidata}. Our method trained with {\it only} RGBD data recovers occluded \hlc[uppink]{empty floors}, \hlc[uppurple]{kitchen cabinets} (row 1) and \hlc[upgreen]{sides of kitchen island}(row 3).} 
    \label{fig:novelwall}
    \vspace{-7mm}
\end{figure}
    
\subsection{Mesh Prediction Results}
\label{sec:exp_meshprediction}

\parnobf{Baselines} Our primary point of comparison is {\bf (1) Mesh DRDF~\cite{kulkarni2022scene}}, which learns to predict the DRDF from direct mesh supervision. For context, we also report the baselines from~\cite{kulkarni2022scene} and summarize them: {\bf (2) {\mpd~\cite{shade1998layered}}} predicts a set of 4 layered depthmaps, where the first represents the depth and the next three represent occluded intersections; {\bf (3) UDF~\cite{chibane2020ndf}} predicts an unsigned scene distance function; {\bf (4) URDF~\cite{chibane2020ndf}} predicts an unsigned {\bf ray} distance function; {\bf (5) ORF} predicts an occupancy function, or whether the surface is within a fixed distance. We note that for each of these approaches, there are a number of variants (e.g., of finding intersections in a URDF along a ray). We report the highest performance reported by~\cite{kulkarni2022scene}, who document extensive and detailed tuning of these baselines.

Our final baseline, {\bf (6) Density Field~\cite{yu2020pixelnerf}} tests the value of predicting a DRDF compared to a density. Like our method, Density Field only uses posed RGBD data for supervision and does not depend on mesh supervision. We adapt pixelNerf~\cite{yu2020pixelnerf} to our setting, modifying the implementation from~\cite{qian2021recognizing} to permit training from a single reference view as input and multiple auxiliary views for supervision. In addition to the standard color loss, we supervised the network to match the ground-truth auxiliary depth RGBD depth as in~\cite{deng2022depth}. At inference time, we integrate along the ray to decode a set of intersections as in~\cite{deng2022depth}.

\begin{table}[t]
    \setlength{\tabcolsep}{3pt} 
    
    \centering 
    \caption{{\bf Matterport3D~\cite{chang2017matterport3d} Acc/Comp/F1Score}. We separate methods that use ground-truth mesh from ones using posed RGBD by a horizontal line. \ours is comparable to the best Mesh supervised method DRDF, and is better than all other mesh based methods on Scene and Ray F1 scores.} \vspace{-2mm}
    \label{tab:scene}
    
    \begin{tabular}{lls@{~}s@{~}sr@{~}r@{~}rr@{~}r@{~}r} \toprule
    & & \multicolumn{3}{s}{Scene } & \multicolumn{3}{r}{Ray } \\ 
    \multicolumn{2}{l}{ Method} & Acc & Cmp & F1 & Acc  & Cmp & F1  \\ 
    \cmidrule(r){1-2}\cmidrule(r){3-5} \cmidrule(r){6-8}
      & \mpd\cite{shade1998layered}  & 66.2 & 72.4 & 67.4 & 13.9 & 42.8 & 19.3  \\ 
     & \udf\cite{chibane2020ndf}  & 58.7 & 76.0 & 64.7 & 15.5 & 23.0 & 16.6  \\ 
     Mesh & \occ  & 73.4 & 69.4 & 69.6 & 26.2 & 20.5 & 21.6 \\ 
     & \urdf\cite{chibane2020ndf}  & 74.5 & 67.1 & 68.7 & 24.9 & 20.6 & 20.7 \\ 
     & DRDF~\cite{kulkarni2022scene}  & 75.4 & 72.0 &  71.9 &  28.4 &  30.0 & 27.3 \\ 
     \midrule
     %\hline
     %\drdf  & 74.3 & 71.4 & 71.0 & 29.9 & 24.6 & 27.0 \\
     %\hline
     \multicolumn{2}{l}{\ours}  & 73.7  & 73.5 & 72.1   & 28.2  & 22.6 & 25.1\\
     \multicolumn{2}{l}{Density Field~\cite{yu2020pixelnerf}} & 45.8 & 80.2 & 57.5 & 24.8 & 14.0 & 17.9 \\
    \bottomrule
    \end{tabular}
    %\end{adjustbox}
    \vspace{-4mm}
    \end{table}

\parnobf{Qualitative Results} We show qualitative results from our method and the baselines in Fig ~\ref{fig:comparison}. Our methods outputs are comparable to the outputs of the mesh based DRDF that is trained with much stronger supervision. The density fields approach struggles to model the occluded hits and has a lots of floating blobs in the empty space. We show additional results from \ours in Fig.~\ref{fig:novelwall} where it recovers occlude \hlc[uppurple]{kitchen cabinets}, \hlc[uppink]{empty floor space} behind kitchen island, \hlc[upgreen]{sides of kitchen island}, and hollow bed.

\parnobf{Quantitative Results} We report results in Table~\ref{tab:scene}. Without access to ground-truth meshes, our approach slightly {\it outperforms} Mesh-based DRDF on Scene F1 and approaches its performance on Ray Occ. Our approach matches it in accuracy (i.e., the fraction of predicted points that are correct) but does worse on completeness. We hypothesize that this is due to mesh-based techniques obtaining supervision in adjacent rooms. Nonetheless, our approach outperforms all other baselines besides mesh-supervised DRDF by a large margin (25.1 vs 21.6 F1). \ours has substantially higher performance compared to the Density Field baseline. This shows that density fields representation has difficulties in learning from a supervision of less views with large occlusions. % Whereas our ray based loss explicit handles the occluded parts of the scene in a structured way.

\subsection{Training on Incomplete Data}
\label{sec:exp_incomplete}

\begin{figure}[t]
    \centering
    \includegraphics[width=0.95\linewidth]{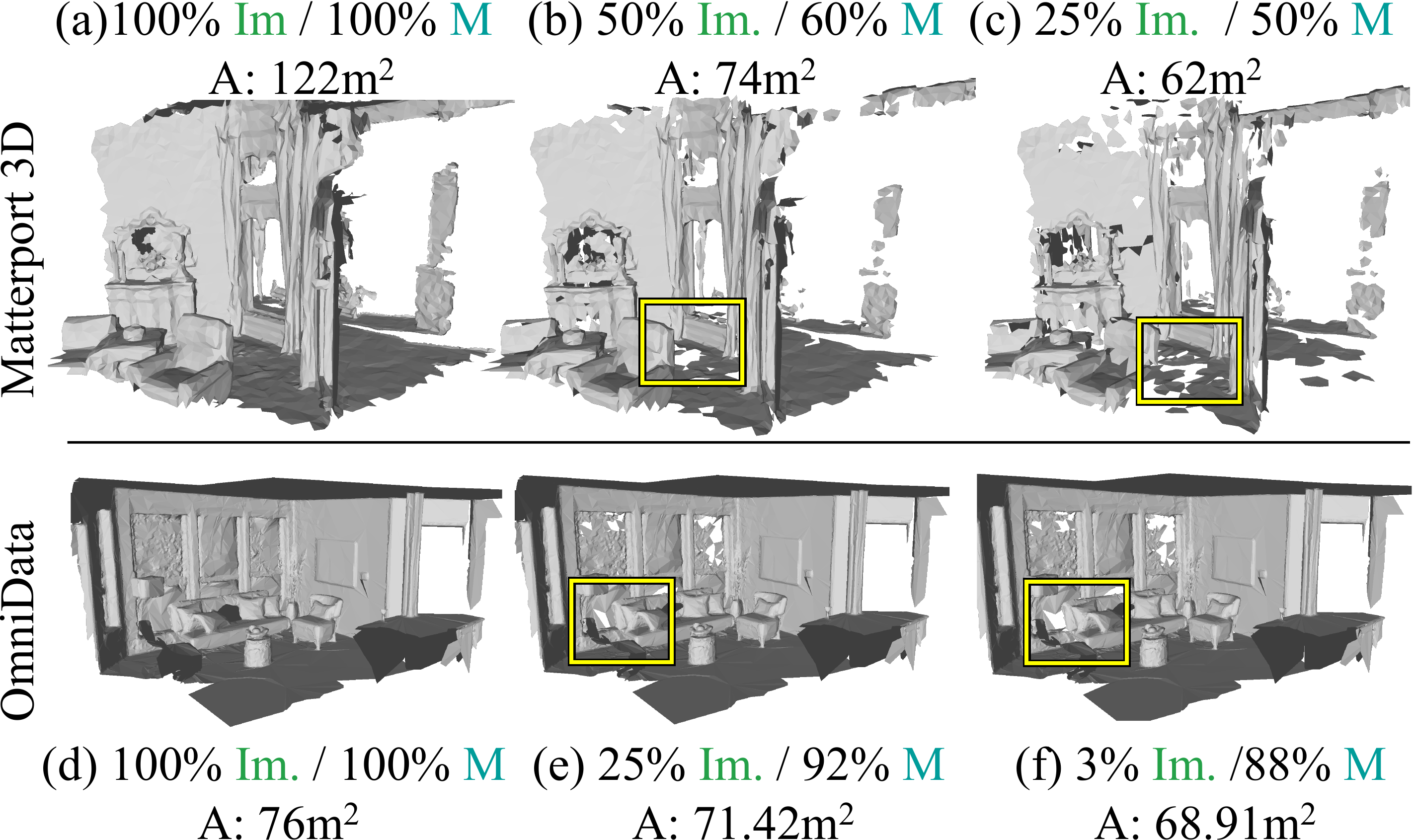}
    \caption{{\bf Mesh Degradation} We show examples of how drop in images creates holes, reduces mesh area available for supervision. In (c), after retaining  ${\approx}\frac{1}{4}$of image data, we only lose ${\approx}\frac{1}{2}$ the mesh area. \imageC: Image Coverage, \meshC: Mesh Coverage, A: Area} 
    \label{fig:decimation}
    \vspace{-6mm}
\end{figure}

One advantage to using posed RGBD data compared to meshes is that it opens the door to learning from more data that is of lower quality. One can use lots of lower-quality scene captures that would produce poor meshes due to substantial amounts of incomplete data. In the case of direct mesh supervision methods, having an incomplete mesh may lead to incorrect signals for a mesh-based system: for instance, the network may learn to predict that an intersection {\it does not} exist simply because it was not scanned. In contrast, for \ours,  missing data simply increases the fraction of points without supervision. We now test this hypothesis by reducing the number of views in datasets.

\parnobf{Optimistic Degradation Setup (ODS)} We simulate the degradation of dataset collection by subsampling views. To avoid conflating errors in training with suboptimal meshing with incomplete data, we {\it optimistically} degrade the meshes to provide an upper bound on supervision. We assume that the mesh with fewer views is identical to the mesh from all views, minus triangles with no vertices in any view. We show ODS mesh examples in Fig.~\ref{fig:decimation}.

We degrade meshes by selecting $1/2^i$ views per dataset for an increasing $i$. While reducing the views linearly impacts the sample count for RGBD training setup, it has a non-linear impact on mesh completeness since a triangle is removed only if {\it all} of the views seeing it are removed (which is unlikely until most views are removed). For any given image retention (\imageC) \%, mesh coverage (\meshC) \% degrades less giving an edge to mesh based methods.

Usually, when dealing with limited data, we use a method called {\it Screened Poisson Reconstruction}(SPR) \cite{kazhdan2006poisson}. However, SPR does not perform well when there is not enough data available. To avoid conflating errors caused by poor quality inadequate meshing, we establish an upper limit on the performance of methods that rely on direct supervision. Our ODS strategy is much better than using SPR, but it cannot be used in real-world scenarios. We employ Open3D's\cite{Zhou2018}'s SPR with hyper-parameters similar to \cite{chang2017matterport3d} to reconstruct meshes. 

\parnobf{Datasets} We evaluate on Matterport3D~\cite{chang2017matterport3d} and apply our method as is without any modifications on OmniData~\cite{eftekhar2021omnidata} which has a substantially different image view distribution, more rooms and more floors  compared to Matterport3D. 

\begin{table}[t]
\centering
\caption{{\bf Robustness to Sparse Data} Performance on partial data on \matterport. We compare ({\bf SPR and ODS}) trained DRDF~\cite{kulkarni2022scene} which
uses mesh supervision and ({\bf Depth}) Depth-based DRDF (ours), which uses posed RGBD supervision. In each row, we degrade the training data and report test performance of the trained model. At 100\% data there is no \meshC degradation for ODS or SPR. Our approach is more robust to drop in \imageC: at 50\% view sparsity (\imageC), models using ODS or SPR suffer substantial performance drops. Scene F1 drop by 16.3 for SPR; 3.5 for ODS; 2.1 for Depth (ours).
}
\label{tab:partialmatterport}
\vspace{-2mm}
\begin{adjustbox}{max width=\linewidth}
\begin{tabular}{l@{~}ls@{~~}s@{~~}sr@{~~}r@{~~}r} \toprule
 & ODS & \multicolumn{3}{s}{Scene F1} & \multicolumn{3}{r}{Ray Occ F1} \\
\imageC\% &  \meshC\% & SPR & ODS  & Depth & SPR & ODS  & Depth \\ 
\cmidrule(r){1-2} \cmidrule(r){3-5} \cmidrule(r){6-8} 
100 & 100 &                71.9  & 71.9                 & 72.1                 & 27.3            & 27.3                  & 25.1 \\
50 & 56 & 55.6  & 68.4 & 70.0  & 21.4  & 23.6  & 24.4 \\
25 & 43 & 56.8   & 66.8 & 70.0  & 21.5 & 21.2  & 24.9 \\ \bottomrule

\end{tabular}
\end{adjustbox}
\vspace{-3mm}
\end{table}

\begin{table}[t]
\centering
\caption{{\bf Robustness to Sparse Data} Performance of ODS (mesh) and Depth (ours) based DRDF on partial Omnidata~\cite{eftekhar2021omnidata} following the same setup as Table~\ref{tab:partialmatterport}. RGBD-based training is substantially more robust to partial data.}
\vspace{-2mm}
\label{tab:partialomnidata}
\begin{adjustbox}{max width=\linewidth}
\begin{tabular}{l@{~}ls@{~~}s@{~~}sr@{~~}r@{~~}r} \toprule
 & ODS & \multicolumn{3}{s}{Scene F1} & \multicolumn{3}{r}{Ray Occ F1} \\
\imageC\% & \meshC\% & SPR & ODS & Depth & SPR & ODS  & Depth \\ 
\cmidrule(r){1-2} \cmidrule(r){3-5} \cmidrule(r){6-8} 
25    & 86 & 63.9 & 77.2                & 72.8            &  26.2     & 40.3                  & 32.1 \\
12.5 & 83 & 62.8 & 75.3    & 70.9 & 26.1  & 37.1.  & 29.3  \\
6.3  & 78 & 40.9  & 73.4  & 71.8  & 5.7  & 32.6  & 28.1  \\
3    & 69 & 42.9  & 69.8 & 70.4  & 3.8  & 20.3  & 26.7  \\ \bottomrule
\end{tabular}
\end{adjustbox}
\vspace{-5mm}
\end{table}

\parnobf{Quantitative Results} We compare models trained on different amount of data available for supervision by using metrics defined in \S\ref{sec:exp_meshprediction}. In Tab.~\ref{tab:partialmatterport} we compare against the Mesh-DRDF trained ODS \& SPR meshes. 
For any given view sampling level, the supervised \meshC\% area is high, resulting in stronger supervision for methods trained with ODS than RGBD. However, on Matterport3D, our method outperforms DRDF on all metrics at 25\% \imageC and outperforms all other baseline methods in Scene F1 at 100\% \imageC.

In Tab.~\ref{tab:partialomnidata} we show robustness trends on OmniData. At 25\% \imageC / 86\% \meshC completion, mesh-based does better. We hypothesize this gain is due to better handling of estimates beyond the room. However, as \imageC reduces mesh degrades, resulting steep fall for ODS DRDF: with 3\% \imageC / 69\% \meshC, MeshDRDF's Ray Occ F1 drops by 20 points; ours is reduced by {\it just} 5.4.
Moreover for SPR DRDF, at low \imageC values, meshing performs dismally and the poor meshing performance translates into poor reconstruction performance: at 6\% \imageC, training DRDF on SPR meshes produces a Ray F1 of just 5.7\%  and a scene F1 of 40.9\%.

\begin{figure}[t]
    \centering
    \includegraphics[width=0.95\linewidth]{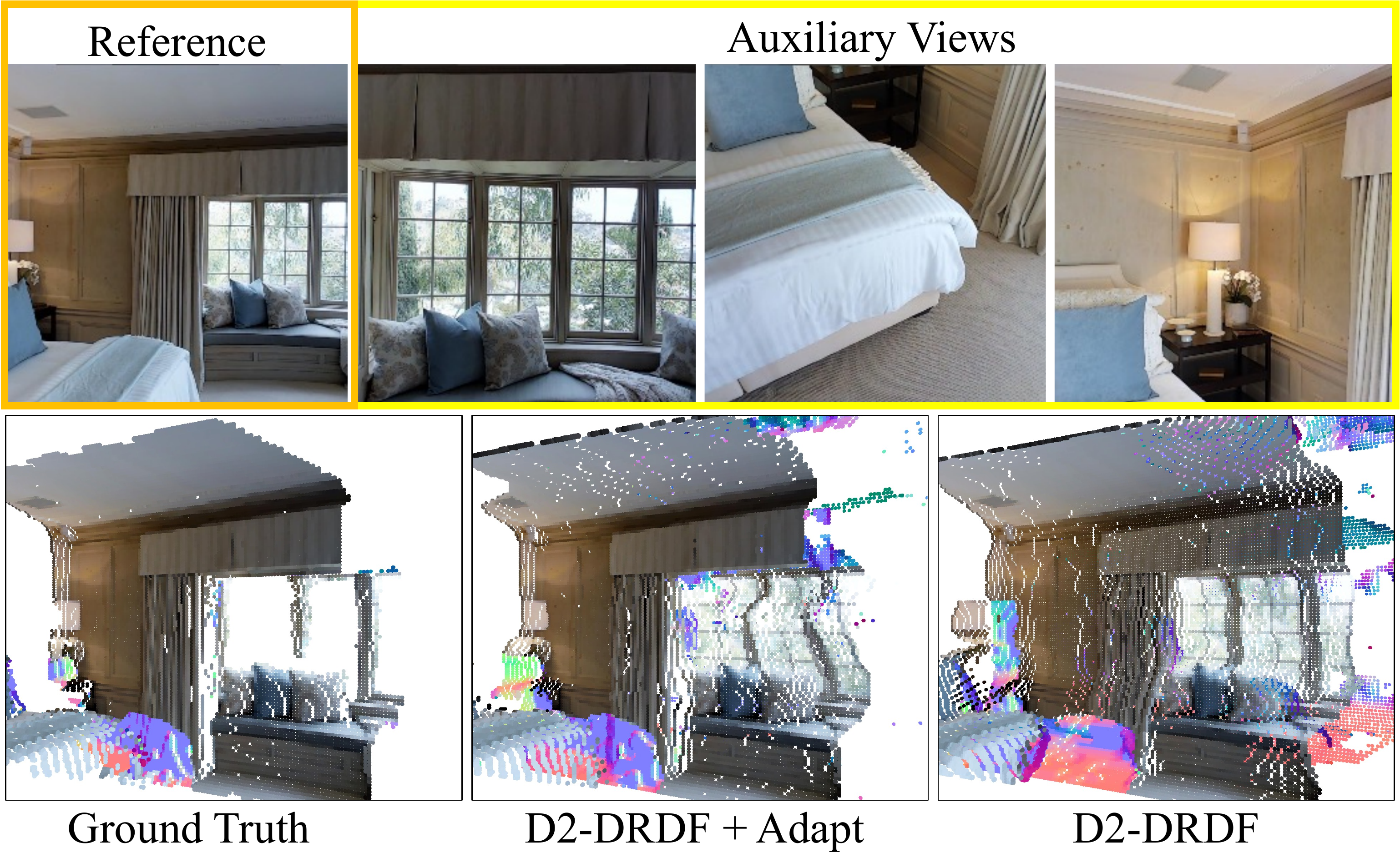}
    \caption{{\bf Adaptation to a few images} We optimize our pre-trained model on the three auxiliary views (yellow) and a reference view (orange). The optimization lets the model  fix reconstructions in occluded regions (\eg bedside wall) as seen \ours + Adapt.} 
    \label{fig:adapt}
    \vspace{-3mm}
\end{figure}
\subsection{Adapting With Multiple Inputs} 
\label{sec:exp_testadapt}

\begin{table}[t]
    \setlength{\tabcolsep}{3pt} 
    \centering 
    % \resizebox{
    %         \width
    %       }{!}{
    %\captionof{table}{{\bf Scene Acc/Comp/F1Score}. \matterport (MP3D) metrics reported @0.5, \tdf metrics reported @0.5, and \scannet reported 0.2. All metrics are higher the better  $\uparrow$. \tsdf outperforms all baselines on F1 demonstrating better scene reconstruction performance. Further discussion under \S\ref{sec:results}}
    \caption{{\bf Scene, Ray F1 Score for Adaptation} Using \ours for adaptation produces better reconstructions that outperform the baselines by a large margin on Ray Occ F1 scores (by 8.9 points). }
    \label{tab:adapt}
    \vspace{-2mm}
    %\begin{adjustbox}{max width=\linewidth}
    \begin{tabular}{ls@{~}s@{~}sr@{~}r@{~}rr@{~}r@{~}r} \toprule
    & \multicolumn{3}{s}{Scene } & \multicolumn{3}{r}{Ray } \\ 
    Method   & Acc & Cmp & F1 & Acc  & Cmp & F1  \\ 
    \cmidrule(r){1-1}\cmidrule(r){2-4} \cmidrule(r){5-7}
     \ours (Full)  & 79.2 & 76.4 & 76.0	& 36.2	& 33.6 &	34.9
     \\
     \ours (Depth) & 78.4 & 70.5 & 	72.5 & 	33.0 &	21.4 &	26.0
     \\
     \ours (Scratch) & 66.4	& 70.3 &	66.0	& 23.7 &	27.4 &	25.4
     \\
     Density Field~\cite{yu2020pixelnerf} & 77.3 &	74.9&	74.6 &	12.8 & 9.9 &	11.2
     \\
    \bottomrule
    \end{tabular}
    %\end{adjustbox}
    \vspace{-7mm}
    \end{table}

Since \ours can {\it directly} train on posed RGBD images, this enables test-time adaptation given a few auxiliary posed RGBD images. We start with the pre-trained model from \S\ref{sec:exp_meshprediction} and then fine-tune for 500 iterations.

\parnobf{Dataset and Metrics} We generate 300 quadruplets of scenes consisting of a reference view as well as {\it three} auxiliary RGBD images with poses. These three auxiliary views are randomly sampled from views that overlap with occluded parts of the reference view (see supp.). We evaluate inferred 3D using the metrics as \S\ref{sec:exp_meshprediction}.

\parnobf{Baselines} The baselines from \S\ref{sec:exp_meshprediction} cannot operate in these settings, since they require meshes for training. Therefore we compare against a number of depth-map-based methods as well as ablations to give context to our results: {\bf (1) Density Field} fine-tunes the density field baseline model from \S\ref{sec:exp_meshprediction}; {\bf (2) Depth-Pretrained} fine-tunes \ours starting with the model from the first stage of training; {\bf (3) Scratch Training} fine-tunes \ours from scratch. 

\parnobf{Results} Our loss and penalty formulations lend themselves well to test-time adaptation as shown in Fig. ~\ref{fig:adapt}. After adaptation, \ours  can resolve uncertainties like the corner of the bed and wall. Table~\ref{tab:adapt} shows \ours does the best.
\subsection*{6. Conclusion}
\noindent
 We presented a method for learning to predict 3D from a single image using implicit functions while requiring only posed RGBD supervision. We believe our method can unlock new avenues with posed RGBD data  becoming available from both consumers as well as robotic agents.

\small{
\parnobf{Acknowledgments}  We thank our colleagues for the wonderful discussions on the project (in alphabetical order).  Richard Higgins, Sarah Jabour, Dandan Shan, Karan Desai, Mohammed  Banani, and Chris Rockwell for feedback.  Shengyi Qian for the help with ViewSeg code used to implement the PixelNeRF baseline. Toyota Research Institute (``TRI”) provided funds to assist the authors with their research but this article solely reflects the opinions and conclusions of its authors and not TRI or any  Toyota entity.
}

% \appendix
%%%%%%%%% ABSTRACT

%%%%%%%%% BODY TEXT

% \tableofcontents

% \clearpage
% \input{sections/extra}

%%%%%%%%% REFERENCES
{\small
\bibliographystyle{ieee_fullname}
\bibliography{egbib}
}
\clearpage
\appendix
\section{Overview}

We discuss crucial details required to implement the results from our paper in \S \ref{sec:implementation}. Then we discuss the under constrained nature of our segment penalties and how multiple DRDFs satisfy them in \S \ref{sec:penalty_functions}. In \S \ref{sec:entropy} we discuss the details of the entropy like loss function followed by additional evaluations and qualitative results in \S \ref{sec:quant}  in \S\ref{sec:qual} respectively. 

\begin{comment}
First in \S\ref{sec:implementation} we discuss additional but crucial details required to implement the results from our paper. Then we follow this up with discussion on the under constrainted nature of our segment penalties and how multiple DRDFs could satisfy them in \S \ref{sec:penalty_functions}. In \S \ref{sec:entropy} we discuss the details of the entropy like loss function followed by additional evaluations and qualitative results in \S{ref:qual} respectively. 
\end{comment}
\section{Implementation Details}
\label{sec:implementation}
We provide additional implementation details for replicating the results in this paper. We will the release code.

\subsection{Data Pre-processing}
We select up to 20 auxiliary views for every reference view from the complete dataset. These auxiliary views are preprocessed along with reference view to create a cached dataset of ray signature to allow faster training. We get supervision for occluded segments of the ray if there is an occluded intersection or if the part of occluded segment is observed from an auxiliary view. Therefore, it is important to sample points from the auxiliary depth map and then convert these points to full rays in the reference view. Such a strategy guarantees that we create rays with more than one OI segment (the first hit) to train our models. After pre-computation using this strategy we have access to a large repository of rays originating from every reference image.

\parnobf{Detecting Events \& Noise in Depth data}
The process of detecting events and creating segments along the ray needs to be robust so as to not allow bad segments. It is critical that we do not allow  erroneous intersections to jump into the ray signature due to missing data. Rays originating from the camera are terminated at $8m$ from the camera as this is maximum extent of the scene we reconstruct (for fair comparison to baselines~\cite{kulkarni2022scene}). We linearly sample $512$ points along this ray and perform event detection for these points. 

Given an auxiliary view ($\pi$) and the a ray, $\rayr$ for every point, $\xB$, we compute the z coordinate in the view frame of the auxiliary view.  We also record the depth map value at the projection $\pi(\xB)$. We sweep along the ray and detect sign changes for difference between the z-coordinate and the recorded depth. A smooth sign change from $+ve$ to $-ve$ or vice versa indicates that there is an intersection at the sign change that is observed from this auxiliary view. However, a sign change with a discontinuity leads to an occlusion event implying the ray entering or exiting the auxiliary view. Since depth data is noisy, for a cluster of missing depth values  we label the start as an exit occlusion event and the end as an entry occlusion event (if the ray becomes visible). 
\begin{figure*}[t]
    \centering
    \includegraphics[width=0.95\linewidth]{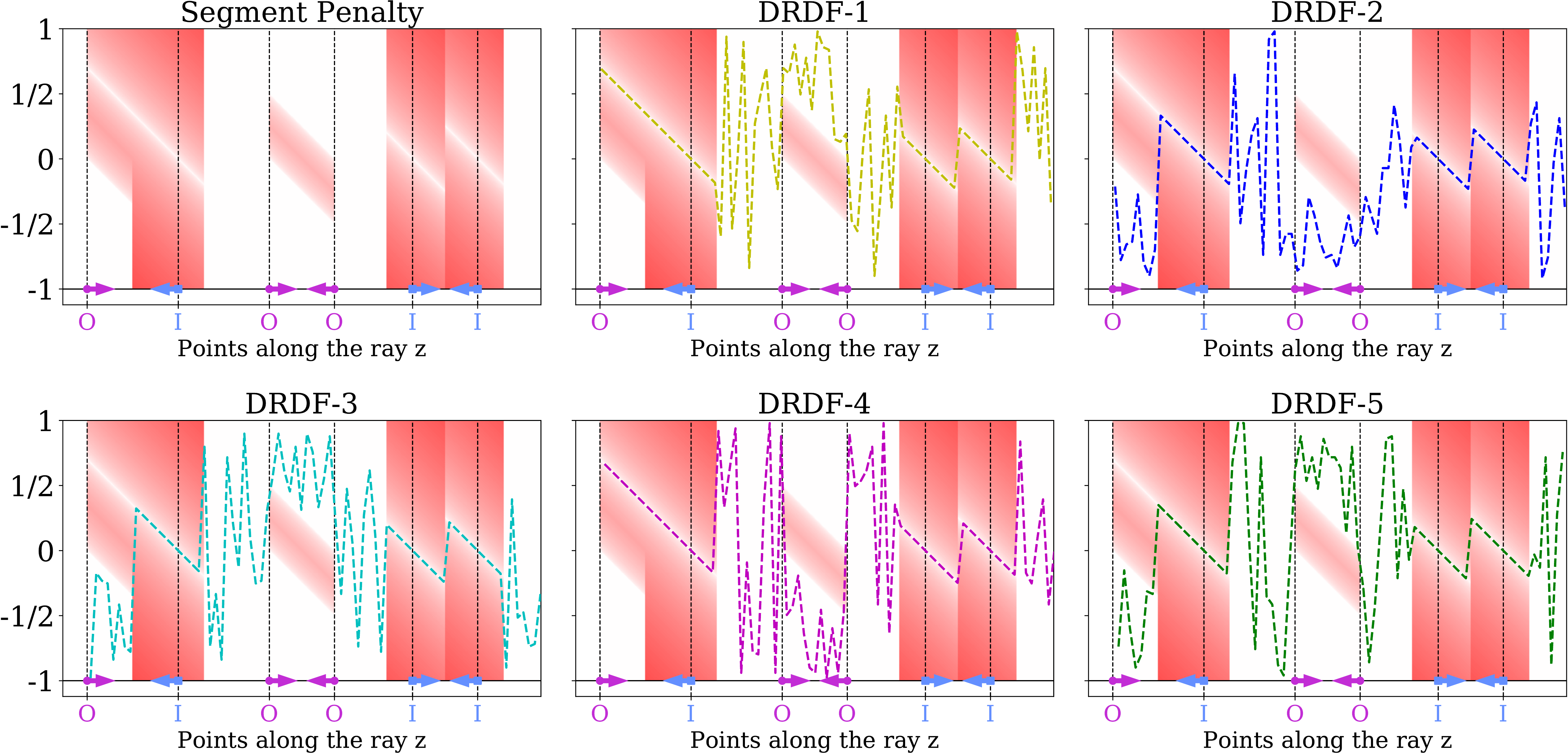}
    \caption{{\bf Under constrained penalty segments }. We consider a  ray that has three segments $\oEvent\iEvent$, $\oEvent\oEvent$, $\iEvent\iEvent$ and goes from {\it left} to {\it right}. On the {\bf first} plot  we show penalty segments for possible DRDF values $\in [-1,1]$ on the Y-axis vs points along the ray $z$. The regions in white have zero penalty and red regions have a high value. . For any particular $z$ these segments give us partial information on the possible values of DRDF for certain parts of the ray. We now show $5$ different possible DRDF functions that all satisfy the penalty plot on top-right. Since some of our segment penalties have inequality constraints there are multiple values possible ($\oEvent\oEvent$). All DRDFs (1-5) match exactly in regions close to an intersection where we have equality constraints (\eg $\iEvent\iEvent$ segment). For regions along the ray not bound by segments DRDF is unconstrained. Penalty legend 0  \includegraphics[width=30pt,height=6pt]{figures/wr_bar.png} 1.7
    }
    \label{fig:penalty}
    \vspace{-4mm}
\end{figure*}
\parnobf{Conflicting segments} 
Detecting events using multiple auxiliary views leads a large set of segments along a particular ray. In order to get an unified segment we drop redundant information \eg we drop an $\oEvent\oEvent$ that is subsumed by an $\iEvent\iEvent$.
Our event detection system is conservative in detecting segments and in case of conflicts we only keep the ray segment that has evidence from most auxiliary views. Since we are training on a large dataset of scene it is convenient to only keep segments that are accurate given the evidence from auxiliary views.

\subsection{Training}
We train our networks in two stages. During the first stage we only train the network to predict the DRDF values until the first intersection. This first stage allows the second stage of the learning process to learn about the occluded scene geometry. Such strategy also lends itself well to use networks bootstrapped on large collection of paired RGB and Depth data such as~\cite{li2018megadepth,Ranftl2020}. We train for half the number of epochs in the first stage and then train on the whole scene including occluded points for the rest half of the epochs (100 + 100).

\parnobf{Architecture}
Our architecture is similar to pixelNerf~\cite{yu2019single, kulkarni2022scene}. We use a pre-trained version of ResNet-34~\cite{he2016deep} from PyTorch~\cite{paszke2019pytorch}, and a 5-Layer MLP with ResNet style skip connections. Our final activation for the last layer is $\tanh$ and it bounds the predicted values in range $[-1,1]$. The MLP takes input the positional embedding ($36$ dimensions) and pyramid of image features at different spatial resolutions ($512$ dimensions). Each of the $5$ hidden layer of our MLP has $1024$ hidden units and the final layer predicts a single scalar value that is the directed ray distance.

\parnobf{Sampling Strategy} In a given reference view with access to only a sparse set of few auxiliary views we do not observe large sections of the occluded geometry. For any given ray in the reference view the predominant segments are the one that are visible in the reference view that capture the first hit. We address this bias in the type of segments by re-balancing the points sampled on all the segments. We randomly sample up to $400$ rays from our preprocessed dataset. For every ray we create linearly spaced $512$ points up to $8m$. We then re-sample points from this large collection of $400 \times 512$ points to keep only $50\%$ points that are visible in the reference view (\ie before the first hit) and $50\%$ points that are occluded from the reference view.

\subsection{Inference}
At inference we consider the input image and a predefined grid in the view frame of size $H \times W \times D$. This grid has $H \times W$ pixel aligned rays. For each ray we decode the ray to a set of intersections along the ray. We speed up parallel decoding for all the rays with the Ray Library~\cite{moritz2018ray}. We use $H=W=D=128$.

\section{D2-DRDF Penalty Functions}
\label{sec:penalty_functions}

\begin{comment}
\begin{figure*}[t]
    \centering
    \begin{adjustbox}{max width=\linewidth}
    \begin{tabular}{c@{\hskip4pt}c@{\hskip1pt}c@{\hskip1pt}c}
    \noindent\includegraphics[width=0.5\linewidth]{supp_figures/penalty.pdf} &
    \noindent\includegraphics[width=0.5\linewidth]{supp_figures/multiple_drdf.pdf}
    \end{tabular}
    \end{adjustbox}
    \caption{{\bf Under Constrained DRDF }. We consider a  ray that has three segments $\oEvent\iEvent$, $\oEvent\oEvent$, $\iEvent\iEvent$ and goes from {\it left} to {\it right}. On the {\bf first} plot  we show penalty segments for possible DRDF values $\in [-1,1]$ on the Y-axis vs points along the ray $z$. The regions in white have zero penalty and red regions have a high value. . For any particular $z$ these segments give us partial information on the possible values of DRDF for certain parts of the ray. We now show $5$ different possible DRDF functions that all satisfy the penalty plot on top-right. Since some of our segment penalties have inequality constraints there are multiple values possible ($\oEvent\oEvent$). All DRDFs match exactly in regions close to an intersection where we have equality constraints. For regions along the ray not bound by any segments DRDFs takes free form value.
    }
    % {\bf (right)} Penalty colormap: 0     \includegraphics[width=30pt,height=6pt]{figures/wr_bar.png} 1.7 } 
    \label{fig:penalty}
    \vspace{-4mm}
\end{figure*}
\end{comment}
The penalty functions discussed in \S \textcolor{red}{4} provide a sparse set of supervisions for our network and now we demonstrate that for a particular ray in a simplified setting. Consider a ray with following segments $\oEvent\iEvent$, $\oEvent\oEvent$, $\iEvent\iEvent$ in order. The $\oEvent$ events provide a weaker constraints as compared to the $\iEvent$. Throughout whenever there is an $\iEvent$ it leads to equality constraints in points on the segment closer to the event, while an $\oEvent$ results in an inequality constraint. In Fig \ref{fig:penalty} (top-left) we show the complete penalty function combined across various segments for points along the ray. We show that $\iEvent$ events lead to a singular solution (shown as a single white line) while $\oEvent$ events lead to multiple regions of zero penalty (white regions). In Fig \ref{fig:penalty} we also show multiple possible DRDFs that satisfy the penalty function but are not the ones we want. It is key to see that all the DRDF (1-5)are exactly the same for points on the ray ($z$) that are closest to the $\iEvent$ while vary largely for points close to the $\oEvent$.

However when we train our network our model observers lot of sample of rays from varied different rays, this access to large scale data encourages a singular DRDF that explains all the events while also being simple when dealing with inequality constraint  (occam's razor). Since neural networks are continuous functions they discourage predictions with high frequency. The key ideas that fuel our approach is not events on a single ray but using data across multiple rays and scenes. Since we use a single neural network to fit to our train data it allows us to learn a bigger set of priors which are useful to create a final DRDF and move away from all possible solutions to these given constraints. Now, in \S\ref{sec:entropy} we discuss another regularizer we use to encourage our network to predict sometimes positive DRDF values and sometimes negative DRDF values for occluded segment.

\section{Entropy-Like Loss}
\label{sec:entropy}
In our loss function we use an entropy loss ($\Lent$) to encourage the predicted DRDFs from segment penalties to behave like DRDFs learnt with mesh supervision. One of the key properties of DRDF is the number of samples that have a negative DRDF value is equal to number of samples that have a positive value. This statistic is only true when we are looking at a dataset of rays. Moreover, since the $\oEvent\oEvent$ imposes an penalty function that discourages changing sign hence reduces diversity, our entropy like loss allows the network to easily adapt to this and jump across large loss values. Below we show some analysis on the behavior of the entropy loss.

Given a prediction $\yB \in \mathbb{R}^n$ over n values, we optimize a differentiable
surrogate for the binary entropy of the {\it signs} of the prediction. This objective
aims to ensure that the predictions are equally negative and positive. Both our ideal objective and surrogate objective can be minimized in the limit as $n \to \infty$ by sampling $\yB$ from a symmetric distribution centered on zero (e.g., a uniform one over $[-1,1]$).

Ideally, we would like to minimize the negative of the entropy of the signs, or
\begin{equation}
\label{eqn:hardentropy}
p \log(p) + (1{-}p) \log(1{-}p) \textrm{~~with~~}
p = \frac{1}{N} \sum_{i=1}^N H(\yB_i),
\end{equation}
where $H(\cdot): \mathbb{R} \to \{0,1\}$ is the Heaviside function mapping
a number to its sign. Equation~\ref{eqn:hardentropy} has a unique
minimum in $p$, namely $\frac{1}{2}$, which is achieved when exactly
half of the components of $\yB$ are positive and half are negative. The requirement
of equal positives and negatives can be satisfied in a large variety of ways.

The Heaviside function is, of course, not differentiable, and so we
use a differentiable surrogate and minimize
\begin{equation}
\label{eqn:softentropy}
p \log(p) + (1{-}p) \log(1{-}p) \textrm{~~with~~}
p = \frac{1}{N} \sum_{i=1}^N \sigma(\yB_i),
\end{equation}
where $\sigma(\cdot): \mathbb{R} \to [0,1]$ is a sigmoid function with a temperature. The
sigmoid functions like a soft sign function where $0$ corresponds to negative values and $1$ to positive values. 
Just like the binary one, Equation~\ref{eqn:softentropy} has a single global minimum in $p$ at $\frac{1}{2}$
and a family of minimums in $\yB$. 

One minimum in $\yB$ is created by generating symmetric values, where each component in $\yB$ has a unique corresponding component with the same magnitude but flipped sign, e.g., if there is a $0.75$m prediction, then there must also be a $-0.75$m prediction. More generally, suppose if $n$ is even, and we order $\yB$ such that $\yB_{i} = -\yB_{n/2+i}$ for all $1 \le i < \frac{n}{2}$. Then
$(\sigma(\yB_{i}) + \sigma(\yB_{n/2+i})) = 1$, and so $p = \frac{1}{N} \sum_{i=1}^{n/2} (\sigma(\yB_{i}) + \sigma(\yB_{n/2+i})) = \frac{1}{2}$. This setting would happen in the limit if the components of $\yB$ were symmetrically distributed over an interval $[-a,a]$ for $a \in \mathbb{R}^+$.

Of course, the surrogate function we minimizes permits other solutions that balance out the right way. For instance, given
on minimizer $\yB$ one can generate another minimizer by adding a $\delta_1$ in one component and adding an appropriate $\delta_2$ in another (i.e., $\yB'_i = \yB_i + \delta_i$). This entails picking $\delta_1$ and $\delta_2$ such that
\begin{equation}
\sigma(\yB_1 + \delta_1) - \sigma(\yB_1) = - \left( \sigma(\yB_2 + \delta_2) - \sigma(\yB_2)\right),
\end{equation}
or that sum remains unmodified. Given a chosen a chosen $\delta_1$, some algebra
reveals that
\begin{equation}
\delta_2 = 
\sigma^{-1}\left( 
    \sigma(\yB_1) - \sigma(\yB_1 + \delta_1) + \sigma(\yB_2) 
\right) - 
\yB_2.
\end{equation} Thus, a whole family of minimizers that do not match pairs of samples or have balanced signs is possible. However, the entropy-like loss is not the only function minimized, and the network must also minimize the data term.

\section{Quantitative Evaluations}
\label{sec:quant}
In addition to F1 scores reported in the main paper we provide the complete results in Table ~\ref{tab:robust_matterport_ods}, ~\ref{tab:robust_taskonomy_ods} for behavior of \ours under different levels of sparse data. Please refer to \S \textcolor{red}{5.2} of the main paper for additional details on evaluation, and dataset creation. With decreasing amount of data, the Mesh DRDF baseline suffers a significant drop in completion (cmp) score on both scene and ray based metrics. This leads to precipitous drop in performance for F1 score (-6.1 points) as compared to \ours which only see a drop of (-0.2 points). We observe similar trends in Table \ref{tab:robust_taskonomy_ods} on OmniData.

As described in the main paper, in practice in sparse view settings we have to leverage mesh reconstruction algorithms like SPR to generated meshes from posed RGBD data. Training methods with this data leads to a subpar performance w.r.t to ODS mesh data. For completeness, we provide the full evaluation in Tab. ~\ref{tab:robust_matterport_spr}, ~\ref{tab:robust_taskonomy_spr} for methods trained with SPR mesh data.

\begin{table}[h]
    \setlength{\tabcolsep}{3pt} 
    \centering 
    % \resizebox{
    %         \width
    %       }{!}{
    %\captionof{table}{{\bf Scene Acc/Comp/F1Score}. \matterport (MP3D) metrics reported @0.5, \tdf metrics reported @0.5, and \scannet reported 0.2. All metrics are higher the better  $\uparrow$. \tsdf outperforms all baselines on F1 demonstrating better scene reconstruction performance. Further discussion under \S\ref{sec:results}}
    \caption{{\bf Matterport3D: Robustness to sparse data 
{\it vs} ODS DRDF}.  Scene Acc/Cmp/F1 and Ray Acc/Cmp/F1 scores on different amounts of Matterport3D dataset. ODS DRDF's Cmp scores drop precipitously result in loss of F1 score. \ours is more stable and robust to amount of data.}
    \begin{adjustbox}{max width=\linewidth}
    
    %\begin{tabular}{lls@{~}s@{~}s@{~}s@{~}s@{~}s@{~}c@{~}r@{~}r@{~}r@{~}r@{~}r@{~}r} \toprule
    \begin{tabular}{llc@{~}c@{~}c@{~}c@{~}c@{~}c@{~}c@{~}c@{~}c@{~}c@{~}c@{~}c@{~}c} \toprule
     & & \multicolumn{6}{s}{Scene} & ~ & \multicolumn{6}{r}{Ray} \\
     & & \multicolumn{3}{s}{\odsDRDF} & \multicolumn{3}{s}{\ours} & &  \multicolumn{3}{r}{\odsDRDF} & \multicolumn{3}{r}{\ours} \\
    \imageC\% & \meshC\% & Acc & Cmp & F1 & Acc  & Cmp & F1 & & Acc & Cmp & F1 & Acc  & Cmp & F1     \\
    \cmidrule(r){1-2}\cmidrule(r){3-5}\cmidrule(r){6-8} \cmidrule(r){10-12}\cmidrule{12-15} 
     100 & 100 & 75.4 & 72 & 71.9 & 73.7 & 73.5 & 72.1 & & 28.4 & 30.0 & 27.3 & 28.2 & 22.6 & 25.1 \\
     50 & 56 & 73.0 & 67.9 & 68.4 & 66.6 & 76.9 & 70.0 & & 28.4 & 20.2 & 23.6 & 22.4 & 26.8 & 24.4 \\
     25 & 43 & 72.1 & 65.6 & 66.8 & 64.1 & 80.5 & 70.0 & & 27.3 & 17.3 & 21.2 & 21.4 & 29.7 & 24.9      \\
     %73.0 & 67.9 & 68.4  & 66.9 & 76.9 & 70.0 \\
     %\\ %72.1 & 65.6 & 66.8 & 64.1 & 80.5 & 70.0 \\
    \bottomrule
    \end{tabular}
    \end{adjustbox}
    \label{tab:robust_matterport_ods}
    \end{table}

\begin{table}[h]
    \setlength{\tabcolsep}{3pt} 
    \centering 
    % \resizebox{
    %         \width
    %       }{!}{
    %\captionof{table}{{\bf Scene Acc/Comp/F1Score}. \matterport (MP3D) metrics reported @0.5, \tdf metrics reported @0.5, and \scannet reported 0.2. All metrics are higher the better  $\uparrow$. \tsdf outperforms all baselines on F1 demonstrating better scene reconstruction performance. Further discussion under \S\ref{sec:results}}
    \caption{{\bf Matterport3D: Robustness to sparse data  {\it vs} SPR DRDF}.  Scene Acc/Cmp/F1 and Ray Acc/Cmp/F1 scores on different amounts of Matterport3D dataset. SPR DRDF's Cmp scores drop precipitously result in loss of F1 score. \ours is more stable and robust to amount of data.}

    \begin{adjustbox}{max width=\linewidth}
    
    %\begin{tabular}{lls@{~}s@{~}s@{~}s@{~}s@{~}s@{~}c@{~}r@{~}r@{~}r@{~}r@{~}r@{~}r} \toprule
    \begin{tabular}{lc@{~}c@{~}c@{~}c@{~}c@{~}c@{~}c@{~}c@{~}c@{~}c@{~}c@{~}c@{~}c} \toprule
     &  \multicolumn{6}{s}{Scene} & ~ & \multicolumn{6}{r}{Ray} \\
     &  \multicolumn{3}{s}{\sprDRDF} & \multicolumn{3}{s}{\ours} & &  \multicolumn{3}{r}{\sprDRDF} & \multicolumn{3}{r}{\ours} \\
     \imageC\%  & Acc & Cmp & F1 & Acc  & Cmp & F1 & & Acc & Cmp & F1 & Acc  & Cmp & F1     \\
    \cmidrule(r){1-1}\cmidrule(r){2-4}\cmidrule(r){5-7} \cmidrule(r){9-11}\cmidrule{11-14} 
     100 & 75.4 & 72 & 71.9 & 73.7 & 73.5 & 72.1 & & 28.4 & 30.0 & 27.3 & 28.2 & 22.6 & 25.1 \\
     50  & 51.2	& 65.9	& 55.6 & 66.6 & 76.9 & 70.0 & & 16.3 &	31.2 &	21.4 & 22.4 & 26.8 & 24.4 \\
     25  & 51.9	& 67.7 & 56.8 & 64.1 & 80.5 & 70.0 & & 16.1	& 32.7 &	21.5 & 21.4 & 29.7 & 24.9      \\
     %73.0 & 67.9 & 68.4  & 66.9 & 76.9 & 70.0 \\
     %\\ %72.1 & 65.6 & 66.8 & 64.1 & 80.5 & 70.0 \\
    \bottomrule
    \end{tabular}
    \end{adjustbox}
    \label{tab:robust_matterport_spr}
    \end{table}

\begin{table}[h]
    \label{tab:robust_taskonomy_ods}
    \setlength{\tabcolsep}{3pt} 
    \centering 
    % \resizebox{
    %         \width
    %       }{!}{
    %\captionof{table}{{\bf Scene Acc/Comp/F1Score}. \matterport (MP3D) metrics reported @0.5, \tdf metrics reported @0.5, and \scannet reported 0.2. All metrics are higher the better  $\uparrow$. \tsdf outperforms all baselines on F1 demonstrating better scene reconstruction performance. Further discussion under \S\ref{sec:results}}
    \caption{{\bf Omnidata: Robustness to sparse data {\it vs.} ODS DRDF}. Scene Acc/Cmp/F1 and Ray Acc/Cmp/F1 metrics on different amounts of Omnidata dataset. ODS DRDF's Cmp scores drop precipitously resulting in loss of Ray F1 scores. \ours is more stable and robust to amount of data.}
   
    \begin{adjustbox}{max width=\linewidth}
    %\begin{tabular}{lls@{~}s@{~}s@{~}s@{~}s@{~}s@{~}c@{~}r@{~}r@{~}r@{~}r@{~}r@{~}r} \toprule
    \begin{tabular}{llc@{~}c@{~}c@{~}c@{~}c@{~}c@{~}c@{~}c@{~}c@{~}c@{~}c@{~}c@{~}c} \toprule
     & & \multicolumn{6}{s}{Scene} & ~ & \multicolumn{6}{r}{Ray} \\
     & & \multicolumn{3}{s}{\odsDRDF} & \multicolumn{3}{s}{\ours} & &  \multicolumn{3}{r}{\odsDRDF} & \multicolumn{3}{r}{\ours} \\
     \imageC\% & \meshC\% & Acc & Cmp & F1 & Acc  & Cmp & F1 & & Acc & Cmp & F1 & Acc  & Cmp & F1     \\
    \cmidrule(r){1-2}\cmidrule(r){3-5}\cmidrule(r){6-8} \cmidrule(r){10-12}\cmidrule{12-15} 
     25 & 86  & 82.9& 74.1&	77.2&	69.2&	79.1&	72.8&	&	43.9&	37.3&	40.3&	29.3&	35.6&	32.1\\
     12.5 & 83 & 81.4 & 72.3 & 75.3 & 65.9 & 79.2 & 70.9 & & 41.8 & 33.4 & 37.1 & 26.3 & 33.0 & 29.3
     \\
     6.3 & 78 & 80.5 & 69.5 & 73.4 & 68.8 & 77.5 & 71.8 & &  40.0 & 27.5 & 32.6 & 27.4 & 28.8 & 28.1 
     \\
     %73.0 & 67.9 & 68.4 & 66.6 & 76.9 & 70.0 & & 28.4 & 20.2 & 23.6 & 22.4 & 26.8 & 24.4 \\
     3 & 69 & 80.5 & 63.7 & 69.8 & 63.7 & 81.1 & 70.4 & & 37.1 & 14.0 & 20.3 & 24.4 & 29.5 & 26.7 
     \\
     %72.1 & 65.6 & 66.8 & 64.1 & 80.5 & 70.0 & & 27.3 & 17.3 & 21.2 & 21.4 & 29.7 & 24.9      \\
     %73.0 & 67.9 & 68.4  & 66.9 & 76.9 & 70.0 \\
     %\\ %72.1 & 65.6 & 66.8 & 64.1 & 80.5 & 70.0 \\
    \bottomrule
    \end{tabular}
    \end{adjustbox}
    \label{tab:robust_taskonomy_ods}
    \end{table}

\begin{table}[h]
    % \label{tab:robust_taskonomy_spr}
    \setlength{\tabcolsep}{3pt} 
    \centering 
    % \resizebox{
    %         \width
    %       }{!}{
    %\captionof{table}{{\bf Scene Acc/Comp/F1Score}. \matterport (MP3D) metrics reported @0.5, \tdf metrics reported @0.5, and \scannet reported 0.2. All metrics are higher the better  $\uparrow$. \tsdf outperforms all baselines on F1 demonstrating better scene reconstruction performance. Further discussion under \S\ref{sec:results}}
    \caption{{\bf Omnidata:  Robustness to sparse data {\it vs.} SPR DRDF}. Scene Acc/Cmp/F1 and Ray Acc/Cmp/F1 metrics on different amounts of Omnidata dataset. SPR DRDF's scores are significantly worse as compared to \ours}
    \label{tab:datapercentscene}
    \begin{adjustbox}{max width=\linewidth}
    %\begin{tabular}{lls@{~}s@{~}s@{~}s@{~}s@{~}s@{~}c@{~}r@{~}r@{~}r@{~}r@{~}r@{~}r} \toprule
    \begin{tabular}{lc@{~}c@{~}c@{~}c@{~}c@{~}c@{~}c@{~}c@{~}c@{~}c@{~}c@{~}c@{~}c} \toprule
     & \multicolumn{6}{s}{Scene} & ~ & \multicolumn{6}{r}{Ray} \\
     &  \multicolumn{3}{s}{\sprDRDF} & \multicolumn{3}{s}{\ours} & &  \multicolumn{3}{r}{\sprDRDF} & \multicolumn{3}{r}{\ours} \\
    \imageC\%  & Acc & Cmp & F1 & Acc  & Cmp & F1 & & Acc & Cmp & F1 & Acc  & Cmp & F1     \\
    \cmidrule(r){1-1}\cmidrule(r){2-4}\cmidrule(r){5-7} \cmidrule(r){9-11}\cmidrule{11-14} 
     25 & 65.9 & 64.4 & 63.9 & 69.2 & 79.1 & 72.88  & &  23.7 & 29.3 & 26.2 & 29.3 & 35.6 & 32.1 \\
    12.5  & 66.1 & 62.3 & 62.8 & 65.9 & 79.2 & 70.9  & &  24.0 & 28.5 & 26.1 & 26.3 & 33.0 & 29.3 \\
    6.3  & 45.9	& 39.2	& 40.9 & 68.8 & 77.5 & 71.8  & &  16.4 &	3.5 &	5.7 & 27.4 & 28.8 & 28.1 
     \\
    3  & 47.6 &	41.5& 42.9 & 63.7 & 81.1 & 70.4 & &  18 &	2.1	& 3.8 & 24.4 & 29.5 & 26.7 
     \\
     \bottomrule
    \end{tabular}
    \end{adjustbox}
     \label{tab:robust_taskonomy_spr}
    \end{table}

\subsection{Mesh Degradation under sparse views}
\parnobf{SPR from sparse views}
All baseline models trained with mesh supervision from {\it Screened Poisson Reconstruction} used only the subset of views with depth maps. We use depth maps at $512 \times 512$ resolution and convert to posed point clouds. Our final point cloud for the scene is a combination of all the view point clouds. We estimate per point normals using Open3D's \cite{Zhou2018} nearest neighbour normal estimation. The estimated normals along with the unprojected point cloud is used as input to the SPR at an oct-tree depth of 9. This is the same standard used in reconstructing houses in Matterport3D\cite{chang2017matterport3d}.

\begin{figure}[h]
    \centering
        \vspace{-1mm}
    \includegraphics[width=\linewidth]{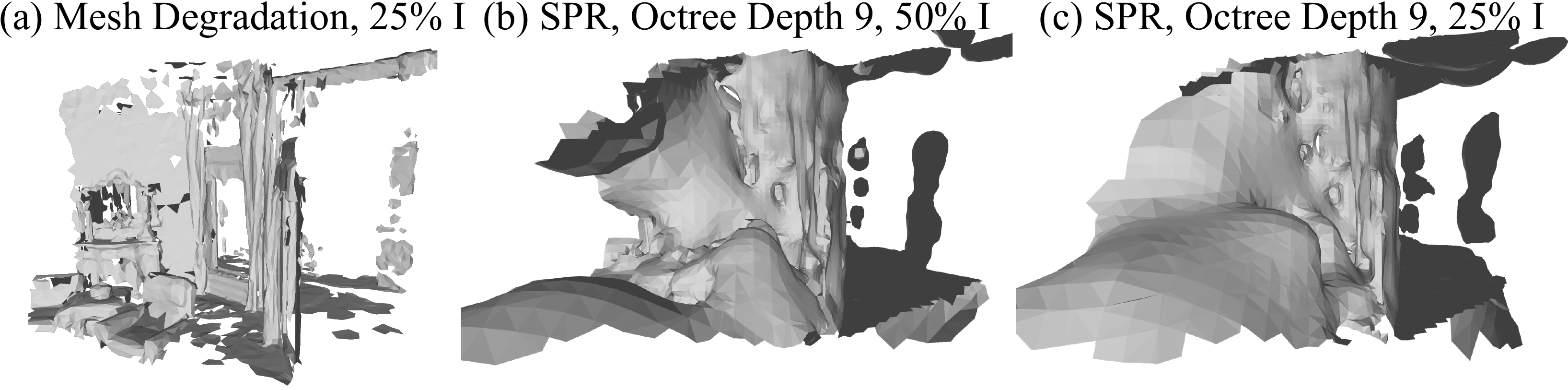}
    \caption{{\bf Screened Poisson Reconstruction (SPR)} (a): degradation with ODS on 25\% image data; (b), (c): SPR reconstructed mesh with 50\% and 25\% image data respectively. The meshes from SPR (b,c) have lots of reconstruction errors and miss on details in the scene whereas the ODS mesh (a) has holes but with reasonable geometry}
    \label{fig:spr_mesh}
\end{figure}

\parnobf{ODS vs. SPR}
Our optimistic degradation setup is an upper bound on the perfromance for methods that train with mesh supervision. In practice a fair comparison would require DRDF baseline trained with meshes reconstructed using {\it Screened Poisson Reconstruction}\cite{kazhdan2006poisson}. Meshes generated using SPR are of much lower quality when compared to ODS meshes. In Fig. ~\ref{fig:spr_mesh} we see SPR reconstrution compared the ODS on Matterport3D at only 25\% image views. The recovered SPR meshes from Open3D's open source implementation~\cite{Zhou2018} have a much lower quality as compared to meshes created by ODS. SPR meshes fail to keep the details of the scene, showcasing that in practice training methods that require mesh supervision is impractical when there are only a sparse set of posed RGBD views.

Overall across both Matteport3D and OmniData we observe that methods trained with SPR mesh achieve a much lower scene F1 and ray F1 scores as compared to same approaches trained with ODS meshes. 

\section{Qualitative Results}

\label{sec:qual}
\begin{figure}[h]
    \centering
    \includegraphics[width=0.35\linewidth]{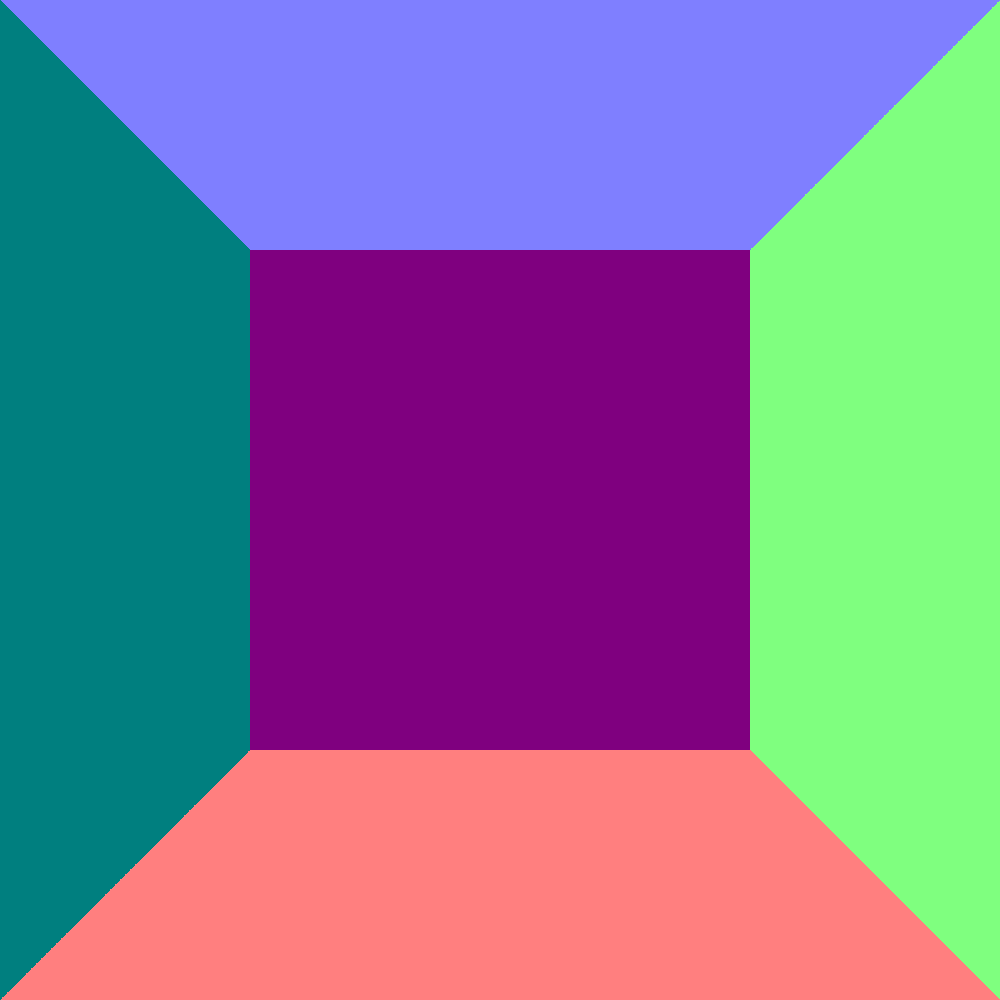}
    \caption{{\bf Surface Normal Legend} We use this surface normal palette to color occluded points reconstructed by all the methods. The surface normals  are computed in the camera frame of reference. In Fig. \ref{fig:matterport_novel_1}, \ref{fig:matterport_novel_2}, \ref{fig:matterport_comp}, \ref{fig:taskonomy_novel_1}, \ref{fig:taskonomy_novel_2} we show reconstructed  \hlc[uppink]{empty floors} are colored in \hlc[uppink]{pink}. The occluded side walls, kitchen cabinets, walls in rooms, other side of kitchen islands are colored in \hlc[upgreen]{green} or \hlc[uppurple]{purple}}
     \label{fig:surface_normal_map}
\end{figure}

\begin{figure*}[t]
    \centering
    \includegraphics[width=0.95\linewidth]{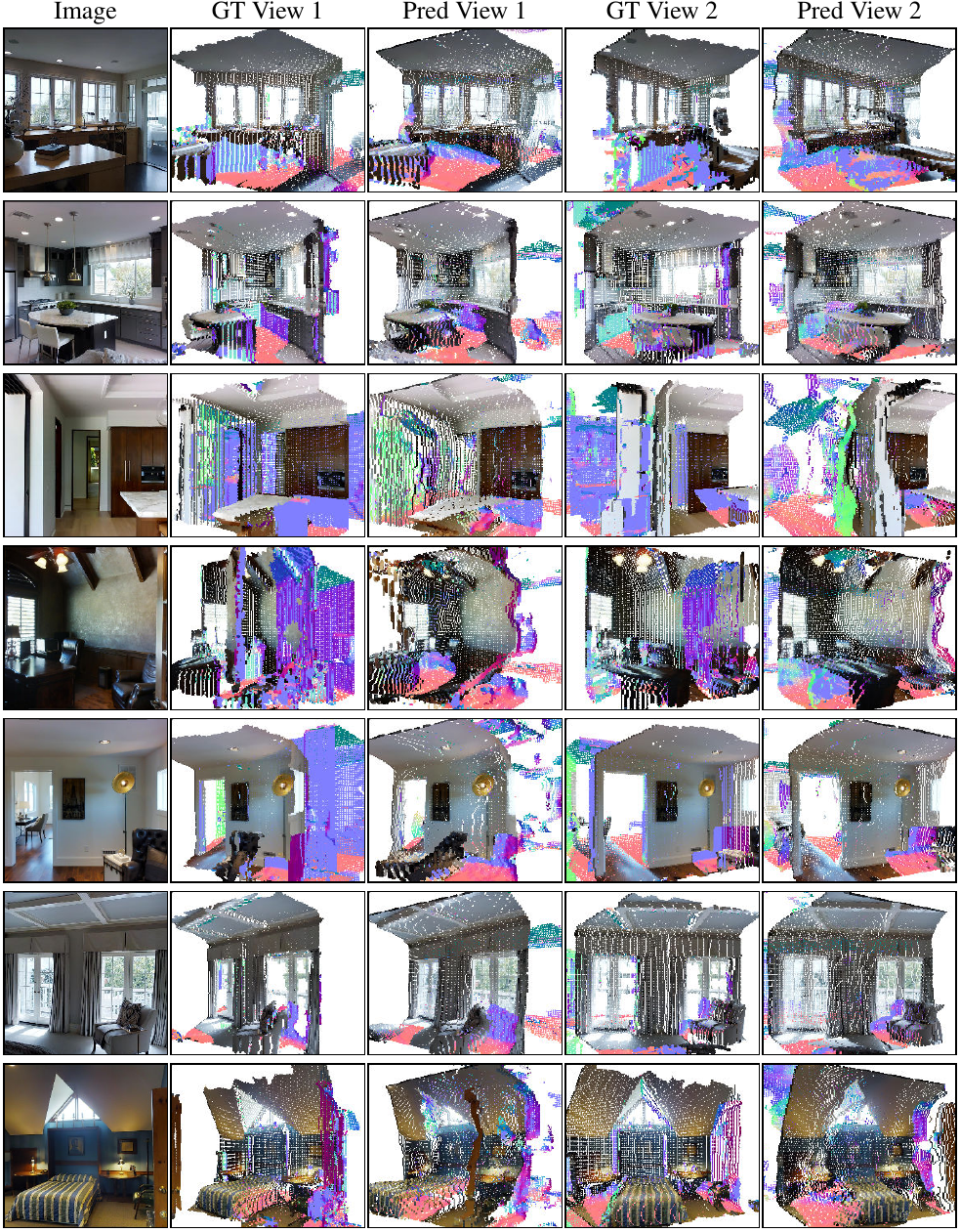}
    \caption{{\bf Matterport3D Novel Views.} We show outputs of \ours  from novel views on previously unseen input images. In column 2, 3 we show ground truth and prediction for view 1 and in 4,5 we show it for view 2. \ours is able to recover the inside of the  \hlc[uppurple]{kitchen island} in {\it rows 1, 2}.  Our model reconstructs the \hlc[uppurple]{occluded wall},  and \hlc[uppink]{empty floor} in {\it rows 5,6}. Please see videos in  \texttt{matterport\_novel.mp4} for additional results}
    \label{fig:matterport_novel_1}
\end{figure*}

\begin{figure*}[t]
   
    \centering
    \includegraphics[width=0.95\linewidth]{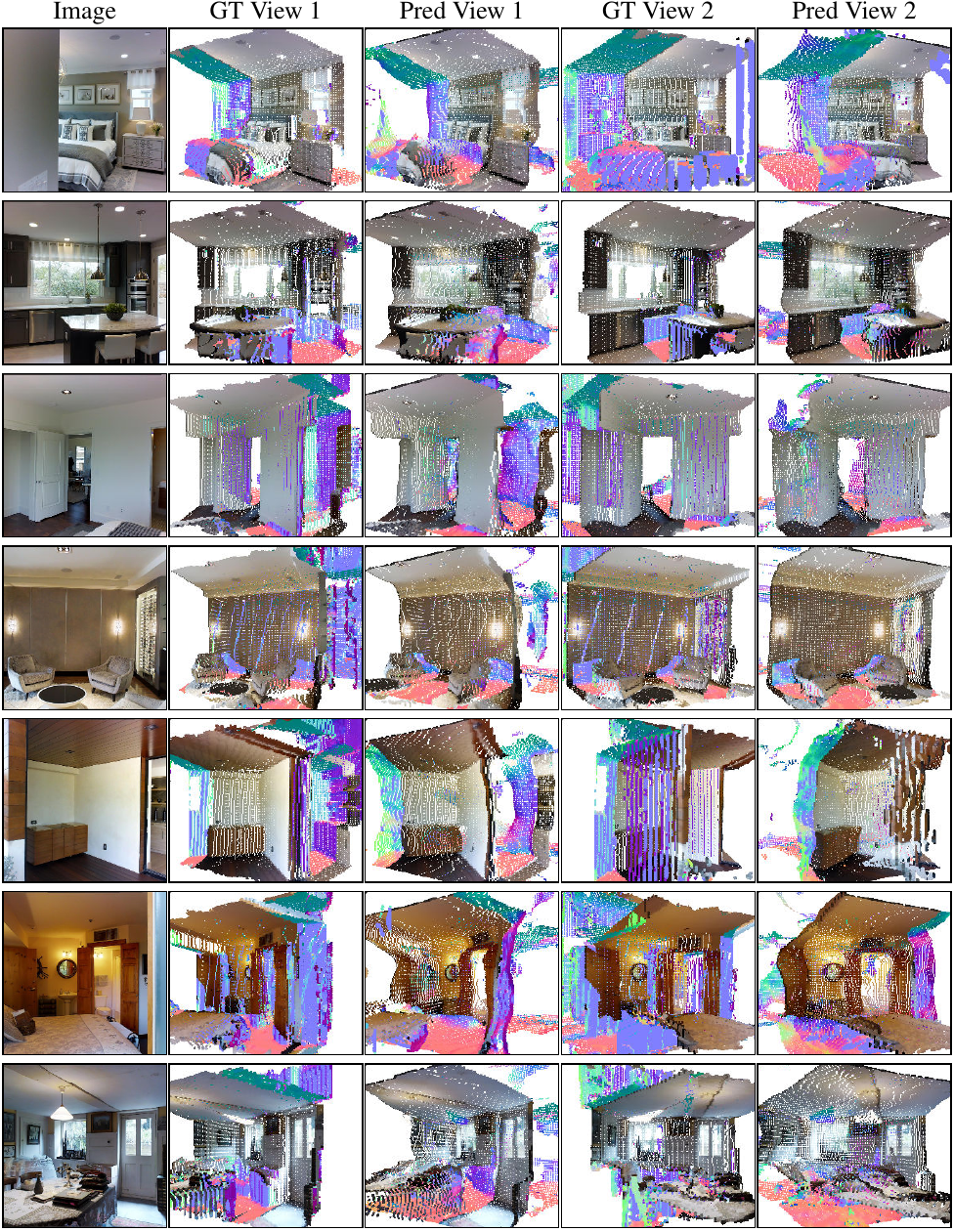}
    \caption{{\bf Matterport3D Novel Views.} Additional results to Fig \ref{fig:matterport_novel_1}. Row 2 shows reconstruction of the occluded \hlc[uppurple]{kitchen island}; Row 3 shows the reconstruction of an occluded empty room. } 
     \label{fig:matterport_novel_2}
\end{figure*}

\begin{figure*}[t]

    \centering
    \includegraphics[width=0.95\linewidth]{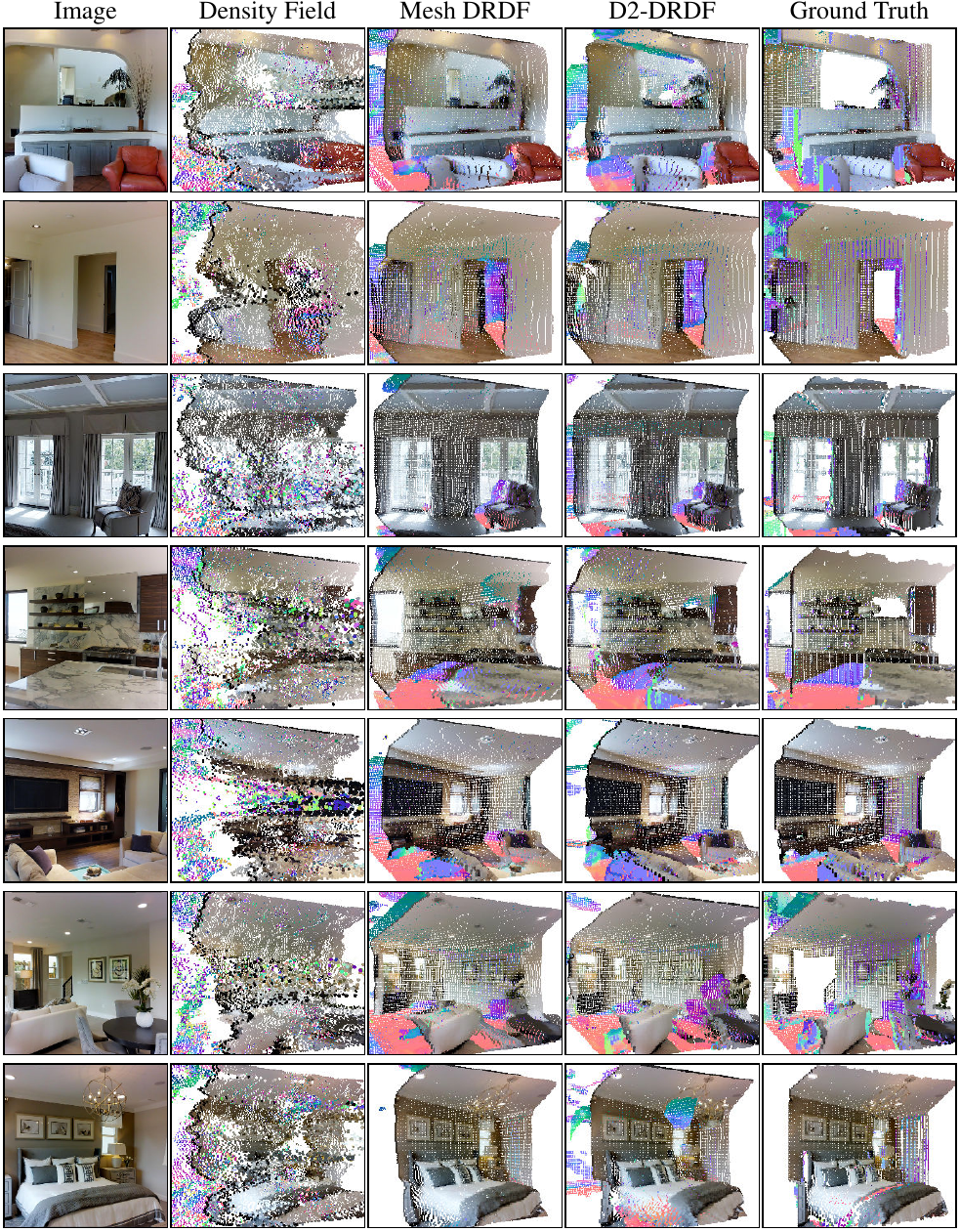}
    \caption{{\bf Matterport3D Baselines.} Column 1 shows the input image for all the methods. Our method (column 4) shows comparable reconstruction results to the mesh supervised DRDF (column 3). The density fields baselines (column 2) fails to recover sharp occluded reconstruction while \ours get occluded parts of floors, kitchen islands, walls, kitchen cabinets. Please see video visualizations} 
      \label{fig:matterport_comp}
\end{figure*}

We show additional qualitative results on Matterport and Omnidata.
\parnobf{Matterport~\cite{chang2017matterport3d} Novel Views} In Fig.~\ref{fig:matterport_novel_1}, ~\ref{fig:matterport_novel_2} we show qualitative outputs of \ours model trained on Matterport\cite{chang2017matterport3d} dataset. We color the occluded reconstructed regions with a surface normal maps from Fig \ref{fig:surface_normal_map}. 
\parnobf{Matterport~\cite{chang2017matterport3d} Comparison to Baselines} In Fig.~\ref{fig:matterport_comp} we show qualitative comparison with baseline methods. The outputs of \ours model trained on Matterport\cite{chang2017matterport3d} are comparable to DRDF model trained with Mesh supervision. The density field baseline trained with posed RGBD data fails at modeling the occluding geometry at test-time. We color the occluded reconstructed regions with a surface normal maps from Fig \ref{fig:surface_normal_map}. 

\begin{figure*}[t]
    
    \centering
    \includegraphics[width=0.95\linewidth]{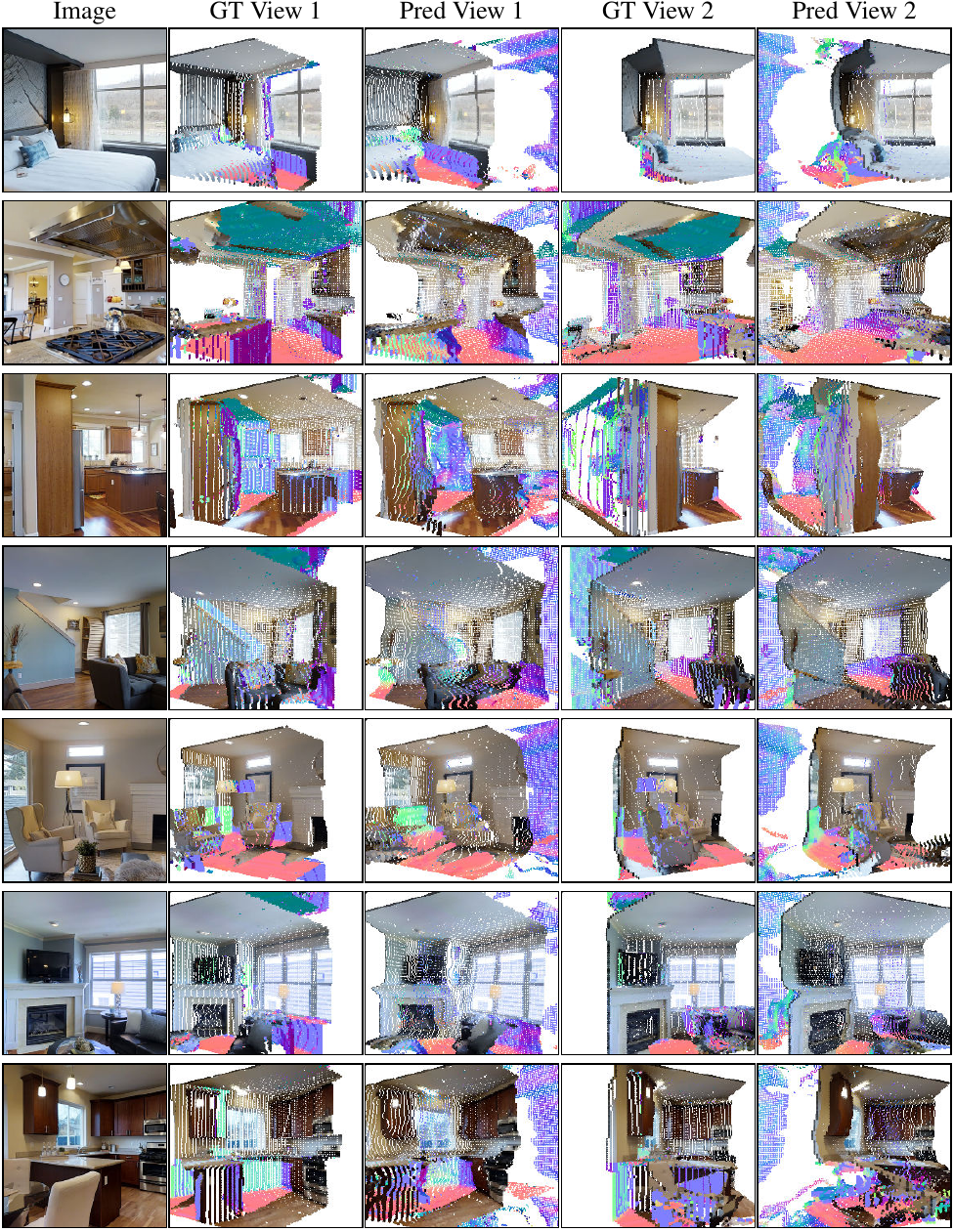}
        \caption{{\bf OmniData~\cite{eftekhar2021omnidata} Novel Views.} We follow the color scheme from Fig \ref{fig:surface_normal_map}, \ref{fig:matterport_novel_1} and show reconstruction results on unseen RGB images from OmniData. Please see the video for additional results} % \texttt{omnidata\_novel.mp4} }
 \label{fig:taskonomy_novel_1}
\end{figure*}

\begin{figure*}[t]
  
    \centering
    \includegraphics[width=0.95\linewidth]{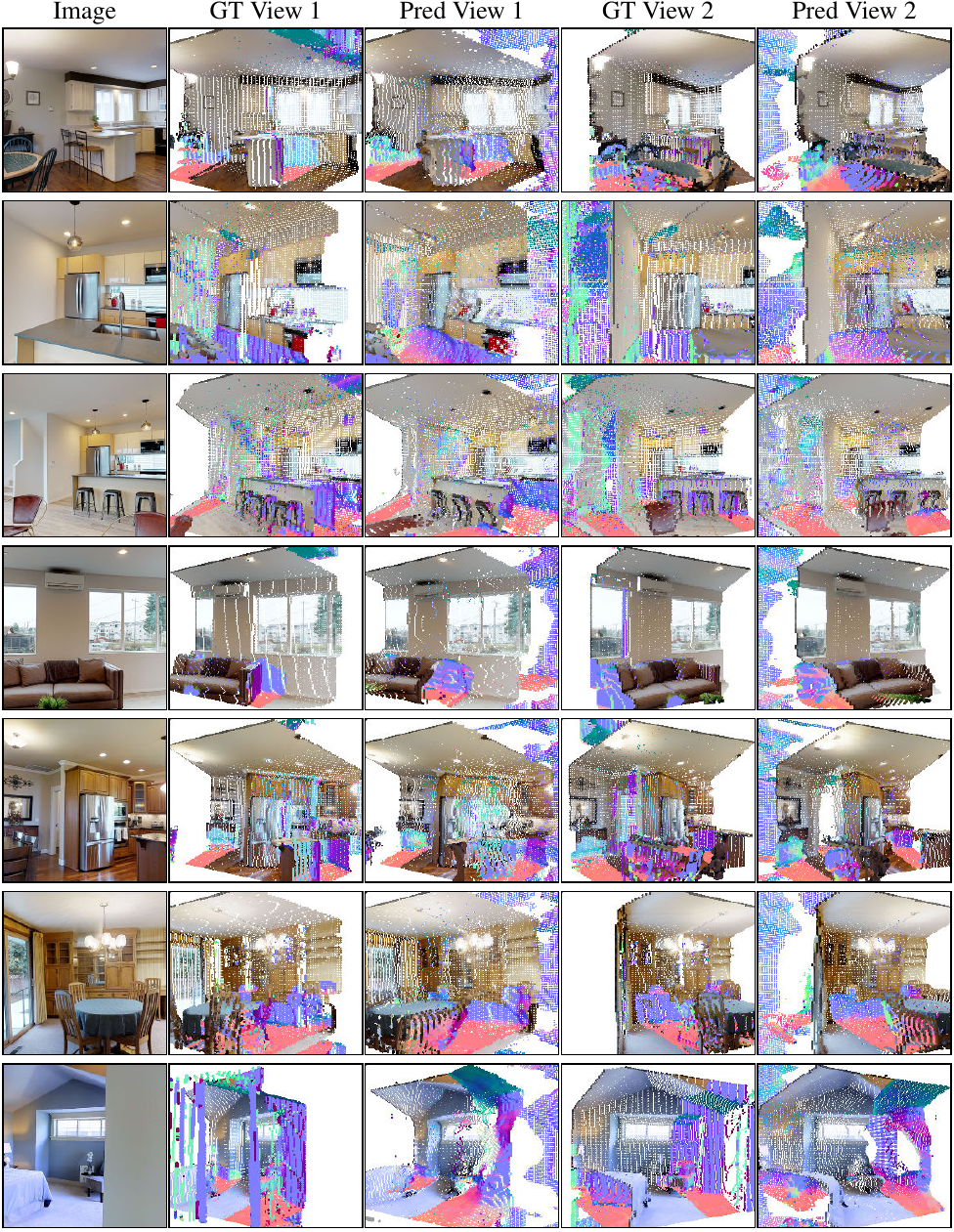}
  \caption{{\bf OmniData~\cite{eftekhar2021omnidata} Novel Views.} We follow the color scheme from Fig \ref{fig:surface_normal_map}, \ref{fig:matterport_novel_1} and show reconstruction results on unseen RGB images from OmniData. Please see the video for additional results } %\texttt{omnidata\_novel.mp4} }
 \label{fig:taskonomy_novel_2}
\end{figure*}

\parnobf{Omnidata~\cite{eftekhar2021omnidata} Comparison to Baselines} In Fig.~\ref{fig:taskonomy_novel_1}, ~\ref{fig:taskonomy_novel_2} we show qualitative outputs from \ours model trained on the OmniData.  We color the occluded reconstructed regions with a surface normal maps from Fig \ref{fig:surface_normal_map}.

% \begin{figure*}[t]
%     \centering
%     \includegraphics[width=0.95\linewidth]{supp_figures/taskonomy_supp_novel_0pt25_p1.pdf}
%     \caption{{\bf OmniData~\cite{eftekhar2021omnidata} Novel Views.} We follow the color scheme from Fig \ref{fig:surface_normal_map}, \ref{fig:matterport_novel_1} and show reconstruction results on unseen RGB images from OmniData.  Please see the video for additional results \texttt{omnidata\_novel.mp4}}
%   \label{fig:taskonomy_novel_1}
% \end{figure*}
% \begin{figure*}[t]
%     \centering
%     \includegraphics[width=0.95\linewidth]{supp_figures/taskonomy_supp_novel_0pt25_p2.pdf}
%     \caption{{\bf OmniData~\cite{eftekhar2021omnidata} Novel Views.} We follow the color scheme from Fig \ref{fig:surface_normal_map}, \ref{fig:matterport_novel_1} and show reconstruction results on unseen RGB images from OmniData. Please see the video for additional results \texttt{omnidata\_novel.mp4} }
%     \label{fig:taskonomy_novel_2}
% \end{figure*}
\end{document}